\documentclass[final]{hubsarxiv}
\graphicspath{{figures/}}

\usepackage{tabularx}
\usepackage{xltabular}
\usepackage{siunitx}
\usetikzlibrary{positioning,arrows.meta,calc}

\title{Where Can a Decision Model Diagnose HVAC Faults? Reasoning Demand, Physical Representation, and Robustness Under Shift}
\shorttitle{Where can a decision model diagnose HVAC faults?}
\shortauthors{Jung}
\author{Wooyoung Jung\affmark{a,}\corrmark\orcid{0000-0003-2746-4983}}
\affil{\affmark{a}Department of Civil and Architectural Engineering and Mechanics, University of Arizona, Tucson, AZ, USA}
\corresponding{wooyoung@arizona.edu}
\paperno{2026-009}
\paperdate{October 2026}
\versionnote{v1 (2026-10-07): first release. v2 (2026-11-05): re-set in the HUBS Lab paper layout; content unchanged.}
\license{CC BY-NC-ND 4.0}
\abstract{Artificial intelligence supports building operations in several forms, each with its own barrier.
Expert rules must be tuned for every system, supervised models need labeled data that buildings
rarely record, and language models return free text that requires human-in-the-loop checking, since
their stated confidence is unreliable. A newer kind of pretrained model, here called a decision
model, returns a probability for every allowed answer, so one model could serve many decisions
without training. This study answers three open questions for fault diagnosis in heating,
ventilation, and air-conditioning systems: which decisions such a model can make, what input it
needs, and whether its probabilities hold when conditions change. On 128 fault days from four public
datasets of real equipment, faults are graded by the reasoning their diagnosis demands, with data
given raw, as physical features, or with Brick topology. The decision model Jev, open language
models, and a supervised model face nine tests that change season, control configuration, or
building. Given physical features, Jev and the larger open model diagnosed faults whose evidence one
feature carries, but not faults that need operating context. Under shift they kept their accuracy
and calibration, while the supervised model lost 0.33 macro-F1 yet led or tied within a building.
Their probabilities still needed correction, and detection was weak. The study maps which faults a
decision model can diagnose and from what input, and supports a division of work in which code
computes the physics and the model ranks candidate faults for an operator.}

\begin{document}
\maketitle

\section{Introduction}\label{sec:intro}
Artificial intelligence (AI) is moving into building operations on different scales and in various
forms~\cite{jung2019hitl}. In fault detection and diagnosis (FDD), for example, expert rules such as
the Air Handling Unit Performance Assessment Rules (APAR) flag faults from fixed relations among
sensor readings \cite{schein2006apar,kim2018review}, and machine learning (ML) classifiers learn
fault signatures from labeled operating data~\cite{zhao2019aifdd}. Beyond diagnosis, model
predictive control (MPC) and reinforcement learning (RL) optimize how equipment is
operated~\cite{drgona2020mpc,wang2020rl}. MPC uses a model of the building to predict its response
over the coming hours and chooses the setpoints that minimize energy use or cost while keeping
occupants comfortable. RL learns a control policy by trial and error, rewarded
for saving energy without sacrificing comfort. The newest forms are language
based~\cite{jung2026prompt,jung2026cot,jung2026buildocc}. Large language models (LLMs) now help
create and query semantic models of buildings from raw point names~\cite{mulayim2024llm}, generate
building energy models from plain-language descriptions~\cite{jiang2024eplusllm}, explain the
decisions of learned controllers to operators~\cite{zhang2024interpretable}, and automate the
development of MPC controllers for heating, ventilation, and air-conditioning (HVAC)
systems~\cite{hu2026autocontrol}. In diagnosis, they mine operating data for faults and
inefficiencies, either alone or guided by domain knowledge~\cite{zhang2024gpt,deng2026graph}.

Each of these forms carries its own burden. Rules must be written and tuned by experts for each
system. Classifiers need labeled data from the system they serve, MPC needs a model of the building,
and RL needs extended interaction with the building or a simulator of it. LLMs need no labels or
building model, but they keep a human in the loop. They produce text, code, or a model, and a
person or another program checks the result before anything acts on the building. The way they
answer keeps them at this distance from the controls. Their output is free text, which software must
parse and validate before acting on it. They can state plausible content that is wrong
\cite{ji2023hallucination}, and the confidence they express in words tends to exceed their accuracy
\cite{xiong2024uncertainty}. Their answers also shift with the wording and order of a prompt
\cite{zhao2021calibrate}, and each answer costs a full text generation, which is slow
and expensive for decisions repeated every few minutes across many units.

A newer kind of model aims closer to the controls. Called here a decision model, it reads a
description of a situation, called the state, together with a set of typed questions about it. Each
question fixes its answer space in advance. It may ask for one choice among named options, a level
on an ordered scale, or a yes or no. The model returns a probability for every allowed answer and
cannot return anything outside that space \cite{typesafe2026jev}. In this it resembles a trained
classifier. But the question and its answers are supplied with each request, so one model can serve
many decisions without being trained on any of them. Jev, released in September 2026, is one such
model. Its developer reports that it was trained so that its stated probabilities track how often it
is right \cite{typesafe2026jev}. Independent evaluations appeared within days of its release
\cite{tang2026audit}. One, on tasks outside buildings, found its probabilities well calibrated when
pooled across tasks but not within a single task \cite{guo2026justaskjev}. Its developer presents it
as a model for fast judgment and advises doing arithmetic in code before the state reaches the
model.

Building automation is full of such decisions. A supervisory controller must judge whether a zone is
occupied~\cite{jung2023occupancy,woo2025infrared}, whether outdoor air is suitable for free cooling,
whether a sensor reading is plausible, and, most consequentially, whether a unit is running normally
or with one of a known set of faults. Each of these decisions has a fixed set of answers, recurs
every few minutes on every unit, and feeds other software. A model that returns a trustworthy
probability for every answer could serve this setting well. The probability could set an alarm
threshold, rank maintenance work by likelihood, or pass an uncertain case to an operator. Before any
such decision is delegated, three things must be known. The first is which decisions these models
can make reliably. The second is what input they need, since a model built for fast judgment may
falter on raw numbers that require calculation. The third is whether their probabilities stay
trustworthy when the season, the equipment, or the building changes.

FDD in HVAC systems is a demanding test of these questions. A training-free model would be valuable
there, because the two established routes stall on the same obstacle. Rules must be tuned by experts
for each system, and classifiers need labeled faults that real buildings rarely record, while
classifiers trained elsewhere must be adapted before use on new equipment
\cite{fan2022transfer,lei2025uda}. A model that diagnoses without training would remove that
obstacle. The task is also hard for such a model, because diagnosis rests on physics. A stuck
outdoor-air damper shows only after the outdoor-air fraction is computed from three temperatures and
compared with the damper command, which is exactly the arithmetic that decision models are advised
to leave to code. FDD also allows the question to be settled with evidence, since public datasets
record faults imposed deliberately on real equipment
\cite{granderson2020data,granderson2023data,ghalamsiah2026data}. Because a wrong diagnosis sends a
technician to the wrong component or leaves a fault running, the trustworthiness of the stated
probabilities matters as much as their accuracy.

Published evaluations of language models for FDD do not yet settle this question. Prompted without
training, LLMs name some faults and miss others \cite{zhang2024gpt,muehlburger2025gemma}, and
fine-tuned models reach high accuracy, but only after training on labeled fault data
\cite{zhang2025domain,langer2024llm}. These studies report accuracy on data from the system and
season used to build or prompt the model. They leave open whether the stated probabilities can be
trusted, how performance changes when the season or the equipment changes, and which kinds of fault
are within reach.

This study addresses three questions.
\begin{itemize}[leftmargin=*,itemsep=1pt,topsep=2pt]
  \item \textbf{RQ1, capability.} Which faults can training-free models diagnose, and how does
    accuracy depend on the reasoning a fault demands?
  \item \textbf{RQ2, representation.} Does computing the physics in code first, as physical
    features, close the gap? Does adding the equipment topology and the latest value of every point
    help further?
  \item \textbf{RQ3, robustness and trust.} When the season, the control configuration, or the
    building changes, do training-free models lose less accuracy than a supervised model? Can their
    stated probabilities be taken at face value?
\end{itemize}

To answer them, this study assembles 128 fault days measured in four real buildings, three built for
research and one occupied, all from public datasets. They cover a rooftop unit, three laboratory
air-handling units (AHUs), and AHUs in an occupied university building. Each fault type is graded by
the reasoning its diagnosis demands, from a single reading that departs from its command to a
pattern that unfolds over time. Every 60-minute window is presented in four forms, ranging from raw
sensor readings to physical features computed in code and the equipment's Brick topology. Jev, open
language models at two sizes, and a rule-based reference diagnose every fault day without training.
A supervised gradient-boosted model joins them in tests that shift the season, the control
configuration, or the building. Accuracy, and how well the stated probabilities match observed
accuracy, are measured in each case.

The study contributes a map of where a training-free decision model can and cannot diagnose HVAC
faults, together with the input it needs and the trust its probabilities deserve when conditions
change. The harmonized splits, labels, and reasoning-level assignments are available from the author
on request.
 
\section{Related work}\label{sec:related}
This section reviews FDD research, which shows what has been achieved and where progress stalls, and
the use of pretrained models for diagnosis, from language models to decision models, including how
operating data are presented to them. It closes with work on when any model's answers can be relied
on, as operating conditions change and as the model states how sure it is.

\subsection{Fault detection and diagnosis: from rules to learned models}\label{sec:related-fdd}
Building FDD methods are commonly grouped into three families
\cite{katipamula2005reviewI,katipamula2005reviewII,kim2018review}. Quantitative model-based methods
compare measurements with the output of a physical model of the equipment. Qualitative,
knowledge-based methods encode expert rules, such as APAR, and need no training data
\cite{schein2006apar}. Process-history methods, also called data-driven methods, learn the normal
and faulty behavior of a system from its operating records. Data-driven work has moved from
statistical monitoring, such as principal component analysis of AHU sensors \cite{wang2004pca},
to machine learning, which learns fault patterns well but needs large amounts of training data
\cite{zhao2019aifdd}. These methods include support
vector machines for chillers \cite{han2011svm}, Bayesian networks for AHUs \cite{zhao2017bayesian},
and gradient-boosted trees for HVAC systems \cite{chakraborty2019xgboost}.

The same limitations recur across this work. Supervised methods reach high accuracy on
laboratory data, but they need labeled faults from the system they serve, and operating buildings
rarely record such labels. Unsupervised methods avoid labels by learning normal behavior and
flagging departures from it \cite{babadi2025unsupervised}. Naming the fault behind a departure,
however, still requires rules, models, or an expert. Results are also hard to compare or reproduce.
An audit of 65 machine-learning FDD studies for HVAC systems found nearly all of them
irreproducible. Of the audited studies, 72\% did not state whether their data were public, and only
two shared code \cite{mukhtar2025reproducibility}. Models that need no training data would avoid the
first of these limitations.

\subsection{Pretrained models for diagnosis}\label{sec:related-models}
Language models are the most recent attempt to diagnose faults without system-specific rules or
labels. Early work gave HVAC operating data to general-purpose LLMs and found them able to name some
faults but unreliable on others \cite{zhang2024gpt}. Fine-tuning on labeled fault data closes most
of that gap. A fine-tuned GPT-3.5 reached near-perfect accuracy on AHU faults from ASHRAE Research
Project 1312 (RP-1312) and outperformed GPT-4 used without fine-tuning \cite{zhang2025domain}.
Domain-specific lightweight models followed \cite{zhang2026lightllm}, and a fine-tuned encoder
model, DistilBERT, reached binary F1 scores of 0.82--0.99 \cite{langer2024llm}. These results depend
on labeled fault data, which real buildings rarely have. Training-free use remains weaker. A
zero-shot model reached a macro-F1 of 0.19--0.25 on real heat-pump data, depending on how the data
were presented, against 0.69 for a supervised XGBoost baseline \cite{muehlburger2025gemma}. Other
work uses LLMs to write detection rules \cite{lin2025pillm}. A review of 66 studies on LLMs for HVAC
operation found none in sustained deployment and recommended LLMs mainly as semantic and workflow
layers \cite{neubauer2026review}. The difficulty is not specific to buildings. Asked which scenario
most likely produced a time series, capable LLMs scored only marginally above random
\cite{merrill2024timeseries}, and in LLM-based forecasters the language model added nothing that a
much simpler component could not provide \cite{tan2024timeseries}. Time-series foundation models,
pretrained on diverse sensor data, generalized only partly on building energy tasks. They forecast
little better than statistical models on unseen data and struggled in complex building environments,
although their representations supported classification \cite{mulayim2025tsfm}. Independent
evaluations of Jev appeared within days of its release in September 2026 \cite{tang2026audit}. Given
accelerometer data for activity recognition without training, it remained far below supervised
models, and adding numerical features lowered its accuracy \cite{konak2026har}. To the author's
knowledge, decision models have not yet been evaluated on fault diagnosis. No study has asked which
kinds of fault are within reach of a model used without training, or why some are and others are
not.

Whether such models succeed depends partly on how operating data are presented to them. On simulated
sensor data from industrial equipment, LLMs did best with summary statistics and worse with raw
readings \cite{lee2025llmfdd}. In buildings, an agent given a Brick knowledge graph and evidence
from a semantic rule engine diagnosed fan-coil faults more accurately, and with fewer false alarms,
than an agent without that evidence \cite{puktalae2026kg}. A workflow combining a semantic metadata
model, a causal graph, and precomputed statistics identified three faults in an environmental
chamber \cite{zou2026kg}. Ontologies such as Brick \cite{balaji2016brick} and the fault-oriented
FDD-ON \cite{chen2026fddon} supply the vocabulary for such inputs, and public building graphs built
on Brick supply examples of it \cite{jung2026buildgraph}. Each of these studies examines one form of
grounding on one system. None compares raw data, computed features, and topology under a common
protocol, or across different kinds of model.

\subsection{Robustness and trust}\label{sec:related-trust}
A model that diagnoses well on one system must still hold up when conditions change, and its
probabilities must mean what they say. Data-driven diagnosis assumes that the data it is applied to
resemble the data it was trained on. In buildings that assumption breaks in several ways. Seasons
change which operating modes occur, and with them how a fault shows. Control sequences differ
between systems of the same type. Each building has its own equipment, sensors, and loads. Transfer
learning and domain adaptation respond by carrying a trained model to a new building
\cite{fan2022transfer,lei2025uda}, or to air handling units whose faults are unlabeled
\cite{ghalamsiah2025uda,ghalamsiah2026graph}, using data from the target system, labeled or not.
Validation schemes that hold out whole parts of the data, such as entire fault types, have been
proposed to estimate how far such models generalize \cite{tan2020stratification}. All of these start
from a trained model. A model used without training has no training data to overfit, yet a change of
season or equipment can still alter how a fault appears in its input. Whether such a model loses
less accuracy than a trained one when conditions change has not been measured.

The second requirement concerns the probabilities themselves. A diagnosis that comes with a
probability is useful only if the probability can be taken at face value. Of the cases a model calls
90\% likely to be a stuck damper, about 90\% should be. In ML this property is called calibration, a
different sense of the word from calibrating a simulation model to measured data. Calibrated
probabilities let an operator set an alarm threshold or rank maintenance work by likelihood. Modern
neural networks tend to be overconfident, and a simple rescaling of their outputs, called
temperature scaling, corrects much of this \cite{guo2017calibration}. Language models add
distortions of their own. Their answers lean toward particular options and positions
\cite{zhao2021calibrate}. Larger models give well-calibrated probabilities on multiple-choice
questions in a suitable format \cite{kadavath2022know}, but the confidence they state in words tends
to be overconfident \cite{xiong2024uncertainty}. Jev was trained specifically for calibrated
decisions. An independent evaluation on alignment benchmarks found its probabilities well calibrated
when pooled across benchmarks (expected calibration error, ECE, of 0.047), but not within a single
benchmark (median ECE 0.168), because its average probability missed each benchmark's base rate
\cite{guo2026justaskjev}. Another found its confidence exceeding its accuracy by 0.21--0.33 on
questions beyond what it knew, a gap that recalibration did not close
\cite{shankaranarayana2026selfknowledge}. Fault data have very different base rates across buildings
and seasons, so for fault diagnosis calibration has to be judged within each dataset and season as
well as pooled.
 
\section{Methodology}\label{sec:method}
\subsection{Study design}\label{sec:design}
\begin{figure}[!htbp]
  \centering
  \includegraphics[width=0.92\textwidth]{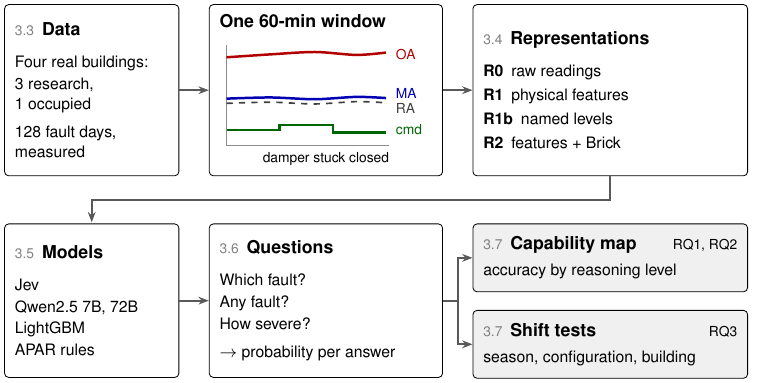}
  \caption{Study design. Numbers in the boxes are the sections that describe each step. Window
    values are illustrative. OA, MA, RA: outdoor-, mixed-, and return-air temperature; cmd: damper
    command.}
  \label{fig:overview}
\end{figure}

The study answers its three questions with two evaluations (Figure~\ref{fig:overview}). The
capability map addresses RQ1 and RQ2. It scores models that need no training on every fault day of
four public datasets, grades each fault type by the reasoning its diagnosis demands, and compares
four ways of presenting the same operating data. The shift tests address RQ3. They hold out a
season, a control configuration, or a building, train a supervised model on the rest, and measure
how much accuracy and probability quality each model loses when conditions change.
Table~\ref{tab:terms} collects the codes and terms used throughout.

\begin{table}[!htbp]
  \caption{Codes and terms used throughout.}
  \label{tab:terms}
  \small
  \begin{tabularx}{\textwidth}
    {@{}>{\raggedright\arraybackslash}p{3.3cm}>{\raggedright\arraybackslash}X
    >{\raggedleft\arraybackslash}p{2.0cm}@{}}
    \toprule
    Code or term & Meaning & Section \\
    \midrule
    L1, L2, L3, L4 & Reasoning levels of a fault type: direct, consistency, context, temporal
      & \ref{sec:taxonomy} \\
    U & Unobservable: the fault leaves no physical trace in the window & \ref{sec:taxonomy} \\
    R0 & Raw readings, a table of 5-minute values & \ref{sec:representations} \\
    R1 & 53 physical features computed in code & \ref{sec:representations} \\
    R1b & R1 with each value replaced by a named level & \ref{sec:representations} \\
    R2 & R1 plus the Brick topology and the latest value of every point
      & \ref{sec:representations} \\
    $\dagger$ & The imposed fault value is written into a logged command, so the fault is visible
      directly in the data & \ref{app:faults} \\
    Capability map & Every fault day scored by models that need no training
      & \ref{sec:evaluations} \\
    Shift test & A test on data from a different season, control configuration, or building than
      the source data & \ref{sec:evaluations} \\
    In-domain reference set & The last 30\% of each source day, sharing that day's conditions
      & \ref{sec:evaluations} \\
    Shifted test set & The data from the held-out season, configuration, or building
      & \ref{sec:evaluations} \\
    Loss under shift & In-domain score minus shifted score & \ref{sec:evaluations} \\
        Macro-F1 & F1 averaged over classes, from 0 to 1 & \ref{sec:metrics} \\
    Top-3 accuracy, mean reciprocal rank & Share of windows whose true class is among the three most
      probable answers; average of one over the true class's rank in the list, from 0 to 1
      & \ref{sec:metrics} \\
    Detection rate, false-alarm rate & Share of fault windows, and of fault-free windows, flagged as
      faulty & \ref{sec:res-detection} \\
    ECE & Expected calibration error, the average gap between stated confidence and observed
      accuracy & \ref{sec:metrics} \\
    AUROC & How well a probability separates fault from fault-free windows over all thresholds,
      from 0.5 (chance) to 1 (perfect) & \ref{sec:res-detection} \\
    \bottomrule
  \end{tabularx}
\end{table}

\subsection{Diagnostic task and reasoning levels}\label{sec:taxonomy}
Each input is a 60-minute window of operating data from one unit. The model returns a probability
for every class in the answer set of that unit's dataset, which contains ``normal'' and each fault
type recorded in the dataset. A diagnosis counts as correct when the most probable class matches the
ground truth. Detection, the narrower question of whether any fault is present, is asked as a
separate yes-or-no question.

Fault types differ in how much reasoning their diagnosis demands from raw measurements. Five levels
capture this difference.
\begin{itemize}[leftmargin=*,itemsep=1pt,topsep=2pt]
  \item \emph{L1, direct.} One measured quantity departs from its command, status, or setpoint. A
    return fan held at a fixed speed shows as a speed signal that stays flat while the supply fan
    modulates.
  \item \emph{L2, consistency.} Two or three measured quantities must satisfy a physical relation. A
    stuck outdoor-air damper is recognized when the outdoor-air fraction implied by the outdoor,
    return, and mixed-air temperatures does not follow the damper command.
  \item \emph{L3, context.} The symptom is masked by control action, appears only in some operating
    modes, or stands out only against an expectation of normal capacity. A biased supply-air
    temperature sensor is typical. The controller drives the biased reading to its setpoint, so the
    sensor itself looks normal, and the fault shows only as a mismatch with the mixed-air
    temperature and the cooling stage.
  \item \emph{L4, temporal.} The symptom is a pattern over time, such as a valve or damper
    oscillating under an unstable control loop.
  \item \emph{U, unobservable.} The fault leaves no physical trace in the window. An example is an
    economizer setpoint error on a day when the outdoor temperature never reaches the affected
    range.
\end{itemize}
The levels follow the kinds of evidence that established diagnostic methods use, ordered by how many
steps separate the raw readings from that evidence. L1 corresponds to limit and command checks, the
simplest knowledge-based diagnosis, which compare one reading with its expected value. L2
corresponds to the mass and energy balances behind model-based methods and rule sets such as APAR,
which combine two or three readings through a physical relation before any comparison is made
\cite{schein2006apar}. L3 adds a condition on the operating state. APAR itself applies different
rules in heating, economizer, and mechanical-cooling modes, and a fault that a control loop
compensates for shows only through its side effects. L4 requires the behavior of a signal over time,
the evidence used to recognize hunting and unstable control. U marks the limit of every method,
since a fault that has changed no measured quantity cannot be detected. Seen from the model's side,
each level adds one operation on numbers: a comparison at L1, an arithmetic combination at L2, a
conditional judgment at L3, and an integration over time at L4. Jev's developer lists arithmetic and
the comparison of dates among its known failure modes \cite{typesafe2026jev}, and language models
reason poorly over time series \cite{merrill2024timeseries}.

A model that diagnoses by recognizing patterns, and not by chaining calculations, should lose
accuracy from L1 to L4. Supplying the L2 relations as precomputed physical features should lift L2
most, since those relations are exactly what the features compute.

\subsection{Data}\label{sec:data}
All data come from real buildings (Table~\ref{tab:datasets}). Three are research buildings built to
test HVAC equipment under controlled conditions, and the fourth is an occupied university building.
In all four, faults were imposed deliberately on the building's own equipment, so every fault day
has a known ground truth. Simulated data are excluded so that no result depends on simulator
fidelity.

\begin{table}[!htbp]
  \caption{Datasets. All four come from real buildings, and all faults were imposed on real
    equipment.}
  \label{tab:datasets}
  \small
  \begin{tabularx}{\textwidth}
    {@{}l>{\raggedright\arraybackslash}Xrr>{\raggedright\arraybackslash}p{2.6cm}@{}}
    \toprule
    Dataset & Equipment, setting, sampling
      & \begin{tabular}[b]{@{}r@{}}Fault days\\(types)\end{tabular}
      & \begin{tabular}[b]{@{}r@{}}Fault-free\\days\end{tabular} & Used for \\
    \midrule
    ORNL RTU & 44\,kW (12.5-ton) rooftop unit, research building, 1 min & 48 (3) & 14
      & Map, seasonal shift \\
    RP-1312 & Paired VAV AHUs, research building, 1 min & 49 (14) & 49
      & Map, seasonal shift, building-shift source \\
    FLEXLAB & One AHU in CAV and VAV modes, research building, 1 min & 21 (6) & 5
      & Map, configuration shift \\
    Nesbitt Hall & AHU-1 and AHU-2, occupied building, 5 min & 10 (3) & 22
      & Map, building-shift target \\
    \bottomrule
  \end{tabularx}
  \par\smallskip\footnotesize ORNL, Oak Ridge National Laboratory; RTU, rooftop unit; AHU,
  air-handling unit; VAV, variable air volume; CAV, constant air volume.
\end{table}

\paragraph{ORNL rooftop unit}
A 44\,kW (12.5-ton) rooftop unit (RTU) at the Flexible Research Platform of Oak Ridge National
Laboratory (ORNL) ran 12 fault cases in each of four seasons between fall 2020 and winter 2022
\cite{granderson2023data,lbnl2022fdd}. The cases are an outdoor-air damper stuck at 5, 10, 50, or
100\% open, an economizer changeover setpoint moved from 10\,\textdegree C to values between 6 and
14\,\textdegree C (from 50 to 42.8--57.2\,\textdegree F), and a supply-air temperature sensor biased
by $\pm$2 or $\pm$4\,\textdegree C ($\pm$3.6 or $\pm$7.2\,\textdegree F). Each case ran for one full
day, giving 48 fault days and 14 fault-free days. Because every case repeats in every season, this
unit gives the cleanest seasonal-shift test of a supervised model.

\paragraph{RP-1312 AHUs}
At the Energy Resource Station (ERS) in Iowa, two matched variable-air-volume (VAV) AHUs ran side by
side in summer 2007, winter 2008, and spring 2008 \cite{wen2011rp1312,ghalamsiah2026data}. One unit
carried an imposed fault while the other ran normally under the same weather and schedule. Each of
the 49 fault days therefore has a fault-free reference from the same day. The 14 fault types include
stuck and leaking dampers and valves, return fan failure and fixed speed, coil fouling, a supply
duct leak, filter blockage, and unstable control loops. Six fault types recur in at least two
seasons, which supports a second seasonal test on the faults that the seasons share.

\paragraph{FLEXLAB AHU}
A single AHU in the FLEXLAB facility of Lawrence Berkeley National Laboratory ran as a
constant-air-volume (CAV) system in March 2017 and as a VAV system in September 2017
\cite{granderson2020data,lin2019oedi}. Its 21 fault days cover stuck and leaking outdoor-air dampers
and coil valves, and 5 days are fault-free. Five fault types occur in both modes, which supports a
test across control configurations.

\paragraph{Nesbitt Hall}
Field data from an occupied university building in Philadelphia cover three AHUs and their VAV
terminals \cite{ghalamsiah2026data}. Faults were imposed by overriding control signals for
windows of 5 to 12 hours. The 10 air-side fault days used here are on AHU-2 (outdoor-air damper
stuck, cooling valve stuck, supply-air temperature bias) and AHU-1 (supply-air temperature bias).
There are 22 fault-free days. Chilled-water plant and schedule faults fall outside this study. Its
AHU-2 faults that also occur in RP-1312, a stuck outdoor-air damper and a stuck cooling valve,
support a test from laboratory units to an occupied building.

\paragraph{Labels and observability}
In the three laboratory datasets every fault was imposed for the whole day, so each minute of a
fault day carries that day's label. Nesbitt labels come from the documented override windows, and
minutes outside a window are not used. Some imposed faults leave no physical trace on some days. An
economizer setpoint moved from 10 to 14\,\textdegree C changes nothing unless the outdoor air
temperature falls between the two values. Each fault day therefore carries an observability flag set
from that day's operating conditions, by rules fixed before any model was run. Unobservable days are
excluded from the capability map and examined on their own.

\paragraph{Harmonization}
The four datasets name and scale their points differently. Every point was mapped to a class of the
Brick schema \cite{balaji2016brick}, an open vocabulary for building equipment and sensors, and to a
common role such as outdoor-air temperature or cooling-valve command. Temperatures are harmonized to
\textdegree F in the data given to the models, and damper and valve signals are in percent open.
This paper reports temperatures in \textdegree C. A small Brick model for each unit links its
equipment to its points. The analysis uses occupied hours only, cut into 60-minute windows at a
15-minute stride.

\paragraph{Reasoning levels in these data}
A level describes the fault as recorded and does not depend on the model or the input format. Each
fault type received the lowest level whose evidence suffices to isolate it from the points recorded
in its dataset. Each fault type was assigned a level from the physics of the fault and the points
available in its dataset, and all assignments were fixed before any model was run. Where the
recorded data contradicted the documentation, the data decided. On the RP-1312 outdoor-air
temperature bias days, the bias acted only inside the controller and the logged temperature is
unbiased. The fault therefore shows only through economizer decisions near the changeover
temperature, and it is placed at L3. Of the 128 fault days in scope, 5 fall at L1, 66 at L2, 52 at
L3, and 5 at L4 (Table~\ref{tab:taxonomy}). \ref{app:faults} lists every fault type with its level,
evidence, and observability condition. Unobservable days are flagged one by one within these levels.
They are analyzed separately, as a test of whether a model withholds confidence where no evidence
exists. The levels are unevenly filled because the public datasets contain few trivial faults and
few oscillation faults. Answer sets range from 4 classes for the ORNL unit to 15 for RP-1312.
Chance-level accuracy is reported for each dataset.

\begin{table}[!htbp]
  \caption{Fault days (fault types) by reasoning level and dataset.}
  \label{tab:taxonomy}
  \small
  \begin{tabular}{lrrrrr}
    \toprule
    Level & ORNL RTU & RP-1312 & FLEXLAB & Nesbitt & Total \\
    \midrule
    L1 & --        & 5 (2)  & --     & --    & 5   \\
    L2 & 16 (1)    & 21 (5) & 21 (6) & 8 (2) & 66  \\
    L3 & 32 (2)    & 18 (6) & --     & 2 (1) & 52  \\
    L4 & --        & 5 (1)  & --     & --    & 5   \\
    \midrule
    Total & 48     & 49     & 21     & 10    & 128 \\
    \bottomrule
  \end{tabular}
\end{table}

\subsection{Representations}\label{sec:representations}
A representation is the form in which a window's data are given to a model. The data are the same in
every form, and only how they are written changes. Each window is written as text in four
representations (Figure~\ref{fig:overview}).
\begin{itemize}[leftmargin=*,itemsep=1pt,topsep=2pt]
  \item \emph{R0, raw.} The unit's core points, resampled to 5-minute steps and rounded to one
    decimal, are sent as a table of time against point. A short header names the equipment, the
    units, the season, and the Brick class of each point.
  \item \emph{R1, physical features.} A library of 53 features (\ref{app:features}) computes in code
    the physical relations a diagnostician would check. Examples are the outdoor-air fraction
    implied by the outdoor, return, and mixed-air temperatures and its difference from the damper
    command, the temperature change across each coil compared with its valve position, the share of
    minutes with a valve commanded closed but a temperature change across its coil, the deviation of
    supply-air temperature from its setpoint, valve and damper reversals per hour, and the APAR rule
    flags \cite{schein2006apar}. Each feature is sent as a name, a value, and a unit. A feature that
    the available points cannot support is left out.
  \item \emph{R1b, named levels.} The R1 features are passed as named levels, such as ``near zero''
    or ``large'', in place of numbers. This follows the provider's advice to convert numbers in code
    before they reach Jev \cite{typesafe2026jev}. The cut points are fixed physical thresholds set
    before any run.
  \item \emph{R2, features and topology.} R1 plus the unit's Brick subgraph, written as short
    statements such as ``AHU hasPart Cooling\_Coil'' and ``Cooling\_Coil hasPoint Valve\_Command''.
    Each point carries its last value in the window. R2 therefore adds two things to R1, the
    topology and the latest value of points that R0 and R1 leave out, such as the return-fan speed
    command.
\end{itemize}
The largest payload is about 1,900 tokens, the word pieces that language models read, which is well
inside every model's input limit. Every model receives the same text for a given window and
representation. \ref{app:questions} shows one window in all three representations.

\subsection{Models}\label{sec:models}
Four models are compared, chosen to separate the kind of model from the input it receives. Jev
represents decision models, and Qwen2.5 at two sizes represents general-purpose
language models, both used without training. LightGBM is a supervised model that learns from labeled
source data, and APAR is a set of expert rules that needs no data at all. Together they span the
forms of AI introduced in Section~\ref{sec:intro}, from expert rules to learned and pretrained
models.
\begin{itemize}[leftmargin=*,itemsep=1pt,topsep=2pt]
  \item \emph{Jev} (TypeSafe AI, version jev-1.13.0) is the decision model introduced in
    Section~\ref{sec:intro} \cite{typesafe2026jev}. Its developer calls it a ``System One'' model
    for fast, structured decisions. It is closed and is accessed through OpenRouter, a commercial
    service that relays requests to model providers, at US\$0.042 per million input tokens at the
    time of the study, with no charge for output. Cost and response times are reported in
    Section~\ref{sec:res-cost}.
  \item \emph{Qwen2.5-Instruct} at 7B and 72B parameters \cite{qwen2024qwen25} are open-weight
    general-purpose language models. They run on a local server with their weights compressed to 4
    bits each so that they fit in memory, served by Ollama, open-source software for running
    language models locally. One model family at two sizes isolates the effect of scale.
  \item \emph{LightGBM} \cite{ke2017lightgbm}, an implementation of gradient-boosted decision trees,
    is the supervised model. Its hyperparameters were fixed before any run and not tuned: 400 trees
    with a learning rate of 0.03, at most 15 leaves per tree and at least 20 windows per leaf, 80\%
    row and column subsampling, light L2 regularization, and class weights balanced by frequency.
    These are conservative settings that favor many small, regularized trees. Tuning was avoided on
    purpose. With one day per fault, the only validation data available are the held-out ends of
    training days, which share each day's conditions. In a pilot fold using raw readings, LightGBM
    recognized economizer faults on 64\% of such windows from days when the fault left no physical
    trace, which is a sign of memorizing day-specific conditions. Tuning on these windows would
    reward that memorization and inflate the in-domain score. Fixed settings also treat LightGBM
    like the training-free models, whose question wording was revised only once. It sees the R0
    table flattened with per-point statistics, the R1 feature vector, or R1 plus a fixed topology
    encoding for R2. The topology encoding is the same for every window of a unit, so LightGBM's
    scores with R2 equal those with R1.
  \item \emph{APAR} \cite{schein2006apar}, applied without training, is the rule-based reference on
    the capability map. A window is flagged as faulty when any applicable rule fires on more than a
    fixed share of its minutes, set before any run. APAR does not name most of the fault types
    studied here, so it is scored for detection only.
\end{itemize}
Open-weight language models were chosen over closed commercial ones because the probability they
assign to every answer label can be read directly, which a calibration comparison
requires. Several commercial models do not expose these probabilities, and confidence stated in
words is a poor substitute \cite{xiong2024uncertainty}. Open weights can also be pinned to an
exact version and remain available, while closed models change or are withdrawn without notice.
Reproducing machine-learning FDD results is already difficult \cite{mukhtar2025reproducibility}.
They run on local hardware at no cost per query, which allows the full set of tests to be repeated
and mirrors how an operator could run a model without sending building data off site. Jev is closed
by necessity, since it is the model under study.

\subsection{Questions, probabilities, and protocol}\label{sec:protocol}

Every window is asked the three questions in Table~\ref{tab:questions}, worded identically for every
model. \ref{app:questions} gives the full wording. Each class carries a one-sentence
description of its physical signature, written from the dataset documentation and identical across
representations. Jev receives all three questions in one request and returns a probability for every
option. The language models receive one question per call and answer with a single word. Each option
is a distinct word (``normal'', ``damper'', and so on), so the first generated word piece identifies
the answer. The probabilities the model gives those words as its next output, read from its 20 most
likely alternatives, are rescaled to sum to one over the answer set. Generation uses temperature 0,
which always picks the most likely word, and a fixed seed. Words replaced letter labels after
letters made the 7B model answer by option position. Its
answer matched itself on 14 of 100 development windows when the option order was reversed, against
95 of 100 with words. The fault probability comes from the yes-or-no question for the training-free
models. For LightGBM it is one minus the probability of normal.

\begin{table}[!htbp]
  \caption{Questions asked of every window, verbatim. The wording was identical for every model.}
  \label{tab:questions}
  \small
  \begin{tabularx}{\textwidth}
    {@{}>{\raggedright\arraybackslash}p{3.6cm}>{\raggedright\arraybackslash}p{2.4cm}
    >{\raggedright\arraybackslash}X>{\raggedright\arraybackslash}p{2.1cm}@{}}
    \toprule
    Question & Answer type & Allowed answers & Used for \\
    \midrule
    ``Which condition best explains this window?'' & One choice among named options
      & ``normal'' and each fault type of the dataset (4--15 options), each shown with its
      description (\ref{app:questions}) & Diagnosis and its calibration \\
    ``Is the unit operating with a fault in this window?'' & Yes or no
      & yes: ``At least one component, sensor, or control setting is not behaving as designed.''
      no: ``Operation is consistent with the design intent for the conditions.'' & Detection \\
    ``How severe is the energy or comfort impact of the condition in this window?''
      & A level on an ordered scale & none, mild, moderate, severe & Severity (secondary) \\
    \bottomrule
  \end{tabularx}
\end{table}

The question wording was tried once before the study, with Qwen2.5 7B on 100 trial windows from the
ORNL unit, and revised once. The revision added a single sentence to the ORNL unit's description,
stating that the unit normally delivers supply air a few degrees above its setpoint. The wording was
then fixed and recorded with a digital fingerprint (a hash) before any scored window was sent, so
that any later change would be detectable. Jev never saw the trial wording. The questions for the
other three datasets follow the same template, with class descriptions taken from their
documentation, and were fixed before any call. The 100 trial windows were not set aside, so they
were also scored later. They make up 2.8\% of the 3,534 ORNL windows on the capability map and fall
in the shifted test sets of three ORNL folds. Removing them changes no capability-map accuracy by
more than 0.004 and no shifted macro-F1 by more than 0.005.

Every request and response is stored, together with the model version that Jev reports. Jev offers
no fixed random seed, and the local server does not decode exactly repeatably, so the same request
can return slightly different answers. Jev and Qwen2.5 7B therefore ran three full passes over one
ORNL split to measure this spread (Section~\ref{sec:res-cost}).

\subsection{Evaluations}\label{sec:evaluations}
On the \emph{capability map}, every in-scope fault day and every fault-free day of the four datasets
is scored without any training. Per-day accuracy is the share of a day's windows whose most probable
class is correct. Unobservable days are scored separately. A model that reads the evidence correctly
should assign them mostly to normal, or spread its probability. Their rate of confident wrong
answers is compared with that on observable days of the same fault type.

Only models that need no training data are placed on the capability map. Most fault types in these
datasets ran on a single day, so a supervised model would have to diagnose faults it never saw in
training. A supervised model is used only where training data for the test faults exist, in the
shift tests described next.

Four \emph{shift tests} compare the training-free models with LightGBM. In each, the model is
tested on data that differ from its source data in one respect, a different season, control
configuration, or building, as happens when a diagnostic tool is moved to new conditions.
\begin{itemize}[leftmargin=*,itemsep=1pt,topsep=2pt]
  \item \emph{Season, ORNL unit.} Each season is held out once, giving four folds over all three
    fault types.
  \item \emph{Season, RP-1312 units.} Each season is held out once, giving three folds, restricted
    to the fault types that the training and test seasons share.
  \item \emph{Control configuration, FLEXLAB unit.} Training uses the VAV mode and testing the CAV
    mode, over the five shared fault types.
  \item \emph{Building, RP-1312 to Nesbitt AHU-2.} The test covers a stuck outdoor-air damper and a
    stuck cooling valve.
\end{itemize}
In every test, LightGBM trains on the first part of each occupied day of the source data. The last
30\% of each such day, after a 60-minute gap, forms an in-domain reference set, and the target data
form the shifted test set. Robustness is the in-domain score minus the shifted score. The in-domain
reference shares each day's conditions with training, so it is optimistic by design. A reference
of whole held-out days was not possible, because each fault intensity ran on a single day and
holding out that day would remove the fault from training. The training-free models answer the same
windows, so every model's scores come from identical inputs.

\paragraph{Sensitivity and ablation}
Four choices are varied one at a time on the test sets, as reported analyses that never select the
main configuration. They are the window length (30, 60, or 120 minutes), the form of the Brick
subgraph (short statements or a nested structure), the R0 time step (1 or 5 minutes), and the order
of answer options for the language models. LightGBM is also refit with learning rates of 0.01 and
0.1 and with 7 and 31 leaves, to show how far its robustness gap depends on the fixed settings.
Jev's answers are also re-elicited with the class descriptions removed or paraphrased, the questions
paraphrased, the unit description removed, irrelevant fields added to the state, and the field order
shuffled. Feature ablations remove one group of R1 features at a time and measure the change in each
model's probabilities, with Shapley additive explanation (SHAP) values \cite{lundberg2017shap} for
LightGBM, which attribute each of its predictions to the input features.

\subsection{Metrics and statistics}\label{sec:metrics}
F1 combines two shares, the share of flagged windows that truly carry the condition and the share
of windows with the condition that are flagged, into one score from 0 to 1. Detection is scored by
F1 at a probability threshold of 0.5, and diagnosis by F1 averaged over classes (macro-F1). Ranking
quality is measured by top-3 accuracy and the mean reciprocal rank, with chance values from the size
of the answer set. Calibration is measured by the ECE, the average gap between stated confidence and
observed accuracy, over 15 bins of equal count, with equal-width bins also reported. Class-wise ECE,
the Brier score (the mean squared error of the probabilities), and reliability diagrams complete the
calibration measures. Predicted severity is compared by the Spearman rank correlation with a level
(none, mild, moderate, or severe) assigned to each fault intensity before any model was run.
Recalibration uses temperature scaling \cite{guo2017calibration} for class probabilities, which
sharpens or flattens them all by one fitted factor, and isotonic regression
\cite{zadrozny2002isotonic} for the fault probability, which maps each stated probability to the
observed fault rate through a fitted increasing curve. Both are fit on the in-domain reference
windows and applied unchanged to the shifted test set. Confidence intervals of 95\% come from
resampling whole days 1,000 times, because windows from one day are not independent. For the loss
under shift, the in-domain and shifted days are resampled separately. The trend across reasoning
levels is the Spearman rank correlation between level and per-day accuracy, with an interval from
the same day resampling. Whether physical features flatten the decline is tested as the difference
in this correlation between R0 and R1.
 
\section{Results}\label{sec:results}
\subsection{Capability map}\label{sec:res-capability}
\begin{figure}[!htbp]
  \centering
  \includegraphics[width=\textwidth]{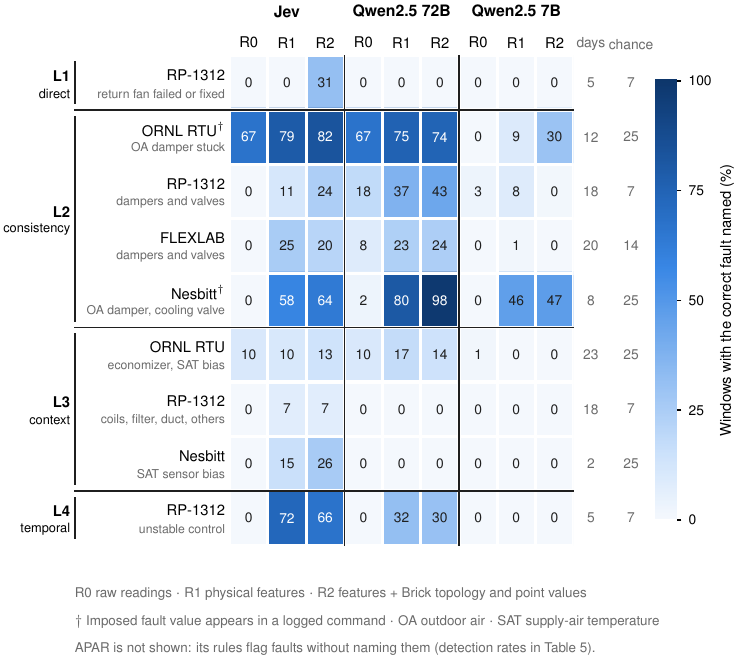}
  \caption{Capability map: share of a fault day's observable windows in which the model named the
    correct fault, averaged over the days in each row (\%). Days counts the fault days with
    observable windows. Chance is one over the size of the dataset's answer set. \ref{app:faults}
    lists every fault type.}
  \label{fig:capability}
\end{figure}

Figure~\ref{fig:capability} shows how often each model named the correct fault, by reasoning level
and dataset. Accuracy is the share of a fault day's observable windows whose most probable answer is
the true fault, averaged over days.

Given raw readings (R0), Jev named the correct fault on 14\% of windows with consistency faults (L2)
on average, pooled over the four datasets, and on almost none at the other levels. Qwen2.5 72B
reached 22\% on consistency faults. For both, most of that accuracy came from one dataset, the ORNL
unit. Qwen2.5 7B named almost no fault in any representation. It answered ``normal'' for most
windows, so its accuracy on fault-free days was high and its accuracy on fault days near zero.

With physical features (R1), Qwen2.5 72B named the correct fault on 46\% of consistency-fault
windows pooled over the four datasets, against 37\% for Jev. It was ahead on RP-1312 (37\% against
11\%) and at Nesbitt Hall (80\% against 58\%). Jev was ahead on temporal faults (L4), 72\% against
32\%, and only Jev named direct faults (L1). On context faults (L3) the two were equally weak, at
9\% pooled. Neither model led at every level.

On the ORNL unit, consistency and context faults ran in every season, and accuracy fell sharply
between them in every representation. With physical features, the Spearman correlation between level
and per-day accuracy was $-0.69$ (95\% CI $-0.90$ to $-0.43$) for Jev and $-0.63$ ($-0.86$ to
$-0.36$) for Qwen2.5 72B. The unit's context faults are an economizer changeover error and a biased
supply-air sensor. Jev named them on 10--13\% of windows, below the 25\% chance level. This
comparison rests on one unit, one consistency fault type, and two context fault types, so it
describes these faults and does not establish a general ordering of the levels.

The decline did not continue to temporal faults. On RP-1312, Jev named unstable control loops on
72\% of windows with features, and Qwen2.5 72B on 32\%, against a chance level of 7\%. On the same
units, Jev's accuracy on consistency and context faults stayed near chance. The features count valve
and damper reversals per hour, which turns an oscillation over time into a single number. Given raw
readings, neither model named any of these faults.

The two direct faults, a failed and a fixed-speed return fan, were never named with features alone,
and were named on 31\% of windows with features and topology (R2). Neither the raw core points nor
the features include the return fan (\ref{app:features}), so neither carried evidence of these
faults. The topology input attaches the last value of every point, including the return-fan speed
command. Qwen2.5 72B named none of these faults in any representation, although the topology input
gave it the same value.

On the ORNL unit and Nesbitt AHU-2, the stuck damper position was written into a logged command, so
the fault appeared directly in the data (marked $\dagger$ in Figure~\ref{fig:capability}). Without
these days, pooled accuracy on consistency faults with features fell from 37\% to 19\% for Jev and
from 46\% to 32\% for Qwen2.5 72B. Qwen2.5 7B's 46\% on Nesbitt's consistency faults came entirely
from such days.

Writing the feature values as named levels (R1b), as Jev's provider advises, lowered the pooled
accuracy on consistency faults from 37\% to 27\% for Jev and from 46\% to 23\% for Qwen2.5 72B.
Jev's accuracy on context and temporal faults changed little, 13\% and 66\% against 9\% and 72\%
with numbers.

Removing one group of features at a time showed which evidence Jev used. On the ORNL winter
windows, removing the mixing and outdoor-air features took its accuracy on consistency faults from
71\% to 0\%, and no other group lowered it by more than 20 points. On the RP-1312 spring windows,
removing the valve and damper reversal counts took its accuracy on temporal faults from 69\% to 0\%.
The same mixing features misled Jev on RP-1312. Without them, its accuracy on consistency faults
there rose from 13\% to 43\%, because it stopped calling every economizer window a damper fault. A
feature that carries the evidence on one unit can thus distract on another.

Jev's inputs were varied on the ORNL winter and RP-1312 spring windows with physical features.
Rewording the three questions changed Jev's most probable answer on 3--6\% of windows and macro-F1
by at most 0.02. Adding 4 or 12 irrelevant fields to the state changed 2--6\% of answers and
macro-F1 by less than 0.01. The class descriptions mattered far more. Given the class names alone,
Jev changed 41\% of its answers on the ORNL unit, and its accuracy on consistency faults fell from
71\% to 6\%. A paraphrase of the descriptions with the same content changed 8--9\% of answers and
raised the ORNL winter accuracy on context faults from 13\% to 43\%. The frozen wording was
therefore not the best available to Jev, and the context-fault accuracies in
Figure~\ref{fig:capability} carry an uncertainty of this size. Removing the unit description lowered
the ORNL accuracy on consistency faults from 71\% to 48\%, and also cut false alarms from 83\% to
54\% of fault-free windows. Shuffling the order of the state fields changed 3\% of ORNL answers,
within Jev's variation between repeated calls, and 13\% of RP-1312 answers. It also made the
yes-or-no answer flag 94\% of the fault-free RP-1312 windows.

A short list of candidates did better than the single most probable answer. With features and
topology, the true fault was among Jev's three most probable answers on 65\% of observable fault
windows, against 46\% by chance, and among Qwen2.5 72B's on 61\%. The gain was uneven. For context
faults, Jev's ranking was no better than chance, so a list did not recover faults the model could
not diagnose. Without the faults whose imposed value appears in a logged command, Jev's top-three
share fell to 58\%, still above the 41\% expected by chance.

\subsection{Detection and false alarms}\label{sec:res-detection}
Detection asks only whether a fault is present. A detector is useful when it flags fault windows
often and fault-free windows rarely. Table~\ref{tab:detection} gives both rates for each dataset.
For Jev and Qwen2.5 they come from a fault probability at a threshold of 0.5, and for APAR from its
rule firings.

\begin{table}[!htbp]
  \caption{Fault detection on all observable fault windows and all fault-free windows (\%). Fault:
    share of fault windows flagged as faulty (detection rate), averaged over fault days.
    Fault-free: share of fault-free windows flagged (false-alarm rate). A useful detector has a high
    first number and a low second one. Language models flag a window when the fault probability is
    at least 0.5. The yes-or-no answer gives that probability directly, and the condition answer
    gives it as one minus the probability of ``normal''.}
  \label{tab:detection}
  \small
  \setlength{\tabcolsep}{3pt}
  \begin{tabular}{@{}lcccccccc@{}}
    \toprule
    & \multicolumn{2}{c}{ORNL unit} & \multicolumn{2}{c}{RP-1312} & \multicolumn{2}{c}{FLEXLAB}
    & \multicolumn{2}{c}{Nesbitt Hall} \\
    \cmidrule(lr){2-3}\cmidrule(lr){4-5}\cmidrule(lr){6-7}\cmidrule(l){8-9}
    Detector & Fault & Fault-free & Fault & Fault-free & Fault & Fault-free & Fault & Fault-free \\
    \midrule
    \multicolumn{9}{@{}l}{\emph{Jev, yes-or-no answer}} \\
    \quad R0 & 48 & 33 & 33 & 17 & 50 & 73 & 10 & 59 \\
    \quad R1 & 95 & 84 & 75 & 57 & 99 & 89 & 88 & 94 \\
    \quad R2 & 83 & 48 & 77 & 55 & 99 & 98 & 90 & 100 \\
    \multicolumn{9}{@{}l}{\emph{Jev, condition answer}} \\
    \quad R1 & 53 & 8 & 63 & 47 & 93 & 60 & 61 & 17 \\
    \multicolumn{9}{@{}l}{\emph{Qwen2.5 72B, yes-or-no answer}} \\
    \quad R1 & 93 & 81 & 95 & 84 & 100 & 100 & 88 & 65 \\
    \multicolumn{9}{@{}l}{\emph{Qwen2.5 7B, yes-or-no answer}} \\
    \quad R1 & 5 & 0 & 11 & 8 & 4 & 10 & 3 & 26 \\
    \midrule
    APAR rules & 88 & 55 & 87 & 75 & 61 & 33 & 90 & 98 \\
    \bottomrule
  \end{tabular}
\end{table}

With physical features, Jev's yes-or-no answer flagged most fault windows, 75--99\% depending on the
dataset. It also flagged 57--94\% of fault-free windows. The two rates were close on every dataset.
With raw readings, Jev flagged fewer windows of both kinds, and at Nesbitt Hall it flagged
fault-free windows more often than fault windows.

On the fault-free days of the ORNL unit, Jev answered ``normal'' to which condition was present for
95\% of windows, yet answered ``yes'' to whether a fault was present for 84\% of the same windows.
Both questions were asked in the same request about the same state. Jev evaluates each question
independently \cite{typesafe2026jev}.

Taking the fault probability from the condition answer cut the false alarms sharply. On the ORNL
unit it flagged 8\% of fault-free windows against 84\% for the yes-or-no answer. Pooled over all
fault-free windows of the four datasets with physical features, it flagged 31\% of fault-free
windows against 76\%, while detecting 65\% of fault windows against 87\%. The area under the
receiver operating characteristic curve (AUROC), which measures how well a probability separates
fault from fault-free windows across all thresholds, was 0.68 for the condition answer and 0.65 for
the yes-or-no answer. Both values lie closer to chance (0.5) than to perfect separation (1). Part of
the separation came from the stuck-damper days whose imposed position appears in logged commands.
Without them, the AUROC of the yes-or-no answer fell from 0.76 to 0.66 on the ORNL unit with
physical features, and at Nesbitt Hall with features and topology (R2) from 0.58 to 0.39, below
chance.

Qwen2.5 7B, in line with its tendency to answer ``normal'', flagged at most 11\% of fault windows.
Qwen2.5 72B erred like Jev, flagging 88--100\% of fault windows and 65--100\% of fault-free
windows. APAR, applied without training, detected 61--90\% of fault windows, but it also raised
alarms on 33--98\% of fault-free windows. At Nesbitt Hall it flagged nearly every fault-free window.

Neither the yes-or-no answers nor APAR could serve as an alarm as configured. False alarms on more
than half of fault-free windows would bury an operator in alerts. Recalibration on source data did
not repair this (Section~\ref{sec:res-calibration}).

\subsection{Robustness under shift}\label{sec:res-shift}
\begin{figure}[!htbp]
  \centering
  \includegraphics[width=\textwidth]{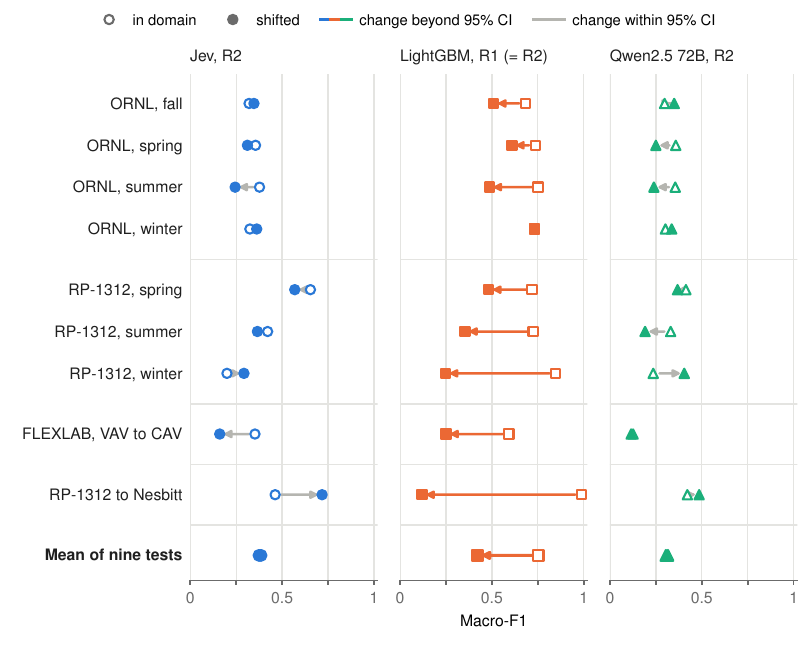}
  \caption{Macro-F1 in domain and on the shifted test set. Each arrow runs from the in-domain score
    to the shifted score, so its length is the loss under shift. Seasons are the held-out season.
    Table~\ref{tab:shift} gives the values and intervals.}
  \label{fig:shift}
\end{figure}

Figure~\ref{fig:shift} shows, for each shift test, the macro-F1 (the F1 score averaged over classes)
on the in-domain reference windows and on the shifted test set. The loss under shift is the first
minus the second, the length of each arrow in the figure. \ref{app:shift} gives both scores with
their confidence intervals.
A training-free model learns nothing from the source data. For Jev and Qwen2.5, a difference between
the two sets therefore reflects only how much harder the target data are to read. For LightGBM, it
also reflects what the model learned from the source data that did not carry over to the target.
A model that learns nothing from the source data has little to lose, so a small loss is not by
itself a strength. For a user, the shifted score is the more informative of the two.
As a floor, a model that answers ``normal'' for every window would score 0.12 on average over the
nine tests, between 0.02 and 0.31 depending on the test, because ``normal'' is the most common
class. On average every model except Qwen2.5 7B scored well above this floor.

Averaged over the nine shift tests, LightGBM's macro-F1 with physical features fell from 0.75 in
domain to 0.42 under shift, a loss of 0.33 (95\% CI 0.28--0.42). The loss ranged from none on the
ORNL winter fold to 0.60 on the RP-1312 winter fold, and reached 0.87 when the building changed. Its
interval excluded zero in every test except ORNL winter. Jev's average loss was 0.00 ($-0.08$ to
0.08) with physical features and 0.01 ($-0.07$ to 0.10) with features and topology (R2), and with
topology its interval included zero in every test. Its shifted score was sometimes higher than its
in-domain score, because a held-out season or building can simply be easier to read. Qwen2.5 72B
lost 0.01 ($-0.06$ to 0.10) with topology, and its interval also included zero in every test. The
intervals rule out an average loss above about 0.10 for either training-free model, but not a
smaller one. Qwen2.5 7B lost almost nothing as well, but from a low level, because it answered
``normal'' for most windows. Measured on the same resampled days, LightGBM lost 0.32 (0.23--0.44)
more than Jev with topology. Its loss was the larger one in all nine tests, and clearly larger on
the RP-1312 summer and winter folds and when the building changed.

On the ORNL unit, LightGBM remained ahead in every held-out season. The difference was clearest in
winter, 0.73 (95\% CI 0.54--0.84) against 0.36 (0.18--0.49) for Jev with topology, and in spring,
0.61 (0.43--0.71) against 0.31 (0.15--0.41). Qwen2.5 72B scored within 0.06 of Jev on every ORNL
fold. On the RP-1312 units and the FLEXLAB unit, both training-free models were comparable to
LightGBM, and their intervals overlapped LightGBM's in every test. In RP-1312 spring, for example,
Jev with topology scored 0.57 (0.32--0.70) and LightGBM 0.48 (0.28--0.60). In no test within one
building did the stability of the training-free models make up for their lower starting point.

The ORNL stuck-damper days carry the imposed position in a logged command, which a trained model can
learn directly. Without these days, every model scored lower on every ORNL fold, because the damper
was the easiest fault for all of them. LightGBM's shifted macro-F1 fell to 0.27--0.50, and Jev's
with topology to 0.11--0.20 on the three remaining classes, close to the 0.11--0.22 of a model that
answers ``normal'' for every window. LightGBM stayed ahead of Jev and Qwen2.5 72B on every fold, by
0.11 to 0.34. Its larger loss under shift also held. With these days and the Nesbitt stuck-damper
days both removed, LightGBM lost 0.30 (0.22--0.40) more than Jev with topology.

Feature attributions for LightGBM (SHAP values) show what it learned. Pooled over the nine tests,
its most influential features were operating conditions that change with the season and the
building, such as the mean return-air and outdoor-air temperatures and the mean damper command,
ahead of the physical relations between readings. For the ORNL stuck damper, the two most
influential were the mean and the spread of the logged damper command.

Ranked by probability, Jev's answers under shift were as useful as LightGBM's. The mean reciprocal
rank, which rewards placing the true class near the top of the list, was 0.70 for both, and the true
class was among the three most probable on 82\% and 88\% of shifted windows, against 67\% by chance.

The order reversed when the building changed. Trained on the RP-1312 units, LightGBM scored 0.99 in
domain and 0.12 (0.06--0.18) on Nesbitt AHU-2, below the 0.31 that answering ``normal'' for every
window would reach. It labeled 82\% of the fault-free Nesbitt windows as a stuck outdoor-air damper
and named the stuck cooling valve on 3 of 73 windows. Jev with topology scored 0.72 (0.48--0.89) and
named the stuck cooling valve on 58 of 73 windows, and Qwen2.5 72B with topology scored 0.49
(0.29--0.66). Part of Jev's score came from the six stuck-damper days, whose imposed damper position
was written into logged damper commands. Without these days, 22 fault-free days and the 2
cooling-valve days remained, and 97\% of windows were fault-free, so answering ``normal'' for every
window would score 0.49. Jev with topology scored 0.48 (0.32--0.61) and Qwen2.5 72B 0.37
(0.21--0.50), no higher than that floor, while LightGBM scored 0.13 (0.07--0.19), far below it. On
this reduced set the reversal therefore reflects LightGBM's false alarms more than any skill of the
training-free models, and with two fault days it rests on little evidence. No model detected faults
better than one that flags every window.

Physical features raised Jev's average shifted score from 0.21 with raw readings to 0.33, and the
topology input, which adds the Brick subgraph and point values, raised it further to 0.37. Jev's
loss under shift stayed near zero in every representation. LightGBM's average loss was 0.35 with raw
readings and 0.33 with features, so the features did not shrink it. The one exception was the ORNL
winter fold, where the loss fell from 0.19 to none. Topology helped Jev most on the RP-1312 units,
raising its spring score from 0.39 to 0.57 and its summer score from 0.21 to 0.37. It added 0.04
when the building changed and lowered the FLEXLAB score from 0.22 to 0.16. Qwen2.5 72B gained almost
nothing from topology, scoring 0.30 on average with features alone and 0.31 with topology.

The comparisons between models held when the window length, the form of the Brick subgraph, or the
time step of the raw readings changed. Windows of 30 and 120 minutes, the nested Brick form, and a
1-minute time step changed Jev's macro-F1 by at most 0.03 in any test, inside every interval of the
main configuration. The 1-minute time step raised Jev's detection F1 by 0.02--0.11. LightGBM's
largest change came with 30-minute windows on the building shift, from 0.12 to 0.30 (0.07--0.43),
still below Jev. Refitting LightGBM with other learning rates and leaf counts changed its shifted
score on the ORNL folds by at most 0.05. Reordering the answer options lowered the open models'
macro-F1 on the ORNL folds and the building shift, by 0.03 on average when the order was reversed
and by 0.03--0.05 when it was shuffled. The largest drop was the 72B's on ORNL winter, from 0.33 to
0.18. Every open-model score therefore carries an uncertainty of about 0.03--0.05 from option order,
as large as the 72B's average gap to Jev with physical features. Jev receives the options as a named
set, and its sensitivity to the order of the state fields is reported in
Section~\ref{sec:res-capability}.

\subsection{Calibration and recalibration}\label{sec:res-calibration}
A model is well calibrated when its stated probability matches how often it is right, so that
answers given with a probability of 0.8 are correct about 80\% of the time. The ECE measures the
average gap. It is reported here for the most probable class, pooled over the nine shift tests.

On the shifted test sets, Jev's ECE was 0.27 (95\% CI 0.21--0.36) with physical features and 0.28
(0.22--0.36) with features and topology (R2), so its stated confidence and its accuracy differed by
more than a quarter on average. Qwen2.5 72B was worse, at 0.43 (0.35--0.50) with topology, and
Qwen2.5 7B worse still, at 0.60 (0.53--0.67). LightGBM reached 0.36 (0.31--0.41). On the capability
map, Jev had the lowest ECE of the training-free models on three of the four datasets. Using 10 bins
in place of 15 changed no pooled value by more than 0.011.

LightGBM's ECE rose by 0.25 (0.18--0.30) from the in-domain to the shifted set. For no
training-free model was the change distinguishable from zero. Jev's ECE rose by 0.01 ($-0.09$ to
0.09) with physical features, and the 72B's fell by 0.02 ($-0.12$ to 0.07) with topology. The Brier
score agreed. LightGBM's rose by 0.49 (0.37--0.61), and every training-free model's change had an
interval containing zero.

Temperature scaling fitted on the in-domain windows lowered the shifted ECE of every training-free
model, from 0.28 to 0.17 for Jev with topology, from 0.43 to 0.16 for the 72B, and from 0.60 to 0.17
for the 7B. Fitted for LightGBM, the same method lowered its shifted ECE only to 0.27, and its rise
under shift stayed at 0.20 (0.15--0.24). Qwen2.5 72B with physical features was the one
training-free case whose rise after recalibration excluded zero, at 0.06 (0.01--0.10). After
recalibration, Jev with topology and the 72B were better calibrated under shift than LightGBM. Jev's
Brier score on the shifted sets was also at or below LightGBM's, 0.62 against 0.70 after
recalibration with overlapping intervals, although LightGBM's macro-F1 was higher. Jev was right
less often but was also confidently wrong less often. Per test (Table~\ref{tab:ece}), the correction
lowered the calibration error on most shifted sets but not all, and it raised Jev's from 0.10 to
0.30 on the RP-1312 summer fold. The ECE computed for each class separately agreed. Only LightGBM's
rose under shift, by 0.13 (0.11--0.16), and recalibration left 0.12 of that rise.

\begin{figure}[!htbp]
  \centering
  \includegraphics[width=\textwidth]{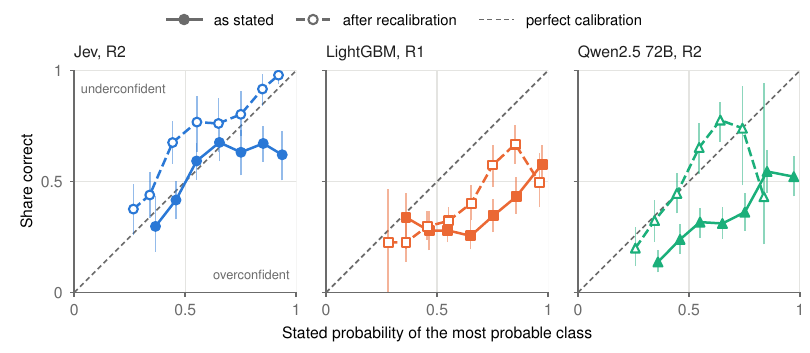}
  \caption{Reliability diagrams on the shifted test sets, pooled over the nine shift tests. Each
    point is a bin of windows grouped by stated probability. Bins with fewer than 30 windows are
    omitted.}
  \label{fig:reliability}
\end{figure}

Figure~\ref{fig:reliability} shows where the errors lay. Above a stated probability of 0.55, Jev's
accuracy hardly changed. Windows it answered with a probability of 0.94 on average were right 62\%
of the time, and those at 0.55 were right 59\% of the time. Recalibration made its probabilities
track accuracy but overshot, so that Jev became underconfident on the shifted sets. LightGBM was
overconfident in nearly every bin, before and after recalibration. Under shift, 59\% of its windows
carried a probability of 0.9 or more, and 58\% of those were right. Qwen2.5 72B was the most
overconfident as stated and came closest to the diagonal after recalibration.

Averaged over the nine shift tests, a detector that flags every window reaches a detection F1 of
0.71. At the 0.5 threshold, LightGBM reached 0.70, Jev 0.67--0.70, Qwen2.5 72B 0.67, and Qwen2.5 7B
0.08. LightGBM's 0.87 on the in-domain set did not carry over. Recalibrating the fault probability
on source data did not help. After the fitted map, the decision at 0.5 followed how common faults
had been in the in-domain set and hardly depended on the window being judged. On the ORNL folds,
where most in-domain windows held a fault, the language models flagged 94--100\% of fault-free
shifted windows. On the RP-1312 folds and the building shift, where faults were rarer, they flagged
0--29\%, and detected correspondingly fewer faults. Their fault probabilities ranked windows too
weakly for a fitted map to do more. After the map, no model's detection F1 rose clearly above that
of flagging every window, 0.71 (0.64--0.75). Jev reached 0.73, inside that interval, and the open
models fell below it.

\subsection{Severity}\label{sec:res-severity}
The severity question, a secondary outcome, asked whether the condition in a window had no, mild,
moderate, or severe impact. Answers were scored against a level assigned to each fault intensity
before any model was run. Predicted and assigned severity correlated weakly across fault days. With
physical features, the Spearman correlation was 0.41 (95\% CI 0.29--0.52) for Jev, 0.42
(0.30--0.52) for Qwen2.5 72B, and 0.23 (0.09--0.35) for Qwen2.5 7B. Within a fault type, predicted
severity rose clearly with the imposed intensity only for the ORNL stuck damper, 0.85 (0.64--0.94)
for Jev, whose position appears in a logged command, and for the 72B on two further fault types.
The severity answer also rarely said ``none'' on fault-free windows, on 26\% of them for Jev with
physical features and on at most 7\% for the open models. It therefore added little to the
diagnosis.

\subsection{Unobservable faults}\label{sec:res-unobservable}
Some imposed faults left no physical trace. Of the fault windows, 1,605 were unobservable, 684 on
days when the fault was visible at other times and 921 on days when it never was. These windows
contain no evidence of the fault, so the reading that matches the data is that nothing is wrong. A
model that reads the evidence should answer ``normal'' or spread its probability over the classes.

With physical features, Jev answered ``normal'' on 88\% of unobservable fault windows, against 49\%
of observable fault windows and 83\% of fault-free windows, and named the imposed fault on 2\%.
Qwen2.5 72B answered ``normal'' on 84\% of unobservable windows and 41\% of observable ones. Both
models therefore told windows with evidence from windows without it. With raw readings, Jev
answered ``normal'' on nearly every window of every kind, so the contrast appeared only with
features.

Neither model lowered its confidence there. Jev gave its most probable answer a probability of 0.7
or more on 63\% of unobservable windows, against 44\% of observable ones. Qwen2.5 72B was more
confident still, with a mean top probability of 0.90 on unobservable windows. Jev's yes-or-no answer
flagged 78\% of unobservable windows, about as many as fault-free windows (76\%). A confident
``normal'' matches the data, yet the equipment was faulty, and nothing in the probability tells an
operator that a fault may be hidden. Whether a fault could show at all has to be judged from the
operating conditions, as in \ref{app:faults}, outside the model.

\subsection{Cost, response time, and repeatability}\label{sec:res-cost}
The study made 190,582 requests to Jev, including every sensitivity test and repeat, for US\$18.69
in total. A request cost about US\$0.0001 and returned in 0.19\,s on average (95th percentile
0.26\,s). Two requests failed. At this price, asking the three questions of one unit every 15
minutes, day and night, would cost about US\$3.40 a year.

On the laboratory server, with two NVIDIA RTX 5000 Ada graphics processors, Qwen2.5 72B took about
2\,s per window for the three questions and Qwen2.5 7B about 0.6\,s. Both are fast enough for a unit
checked every 15 minutes, and neither sends building data off site.

Jev offers no fixed random seed, and the local server's decoding is not exactly repeatable. Three
full passes over the ORNL winter split gave the same most probable answer on 97--99\% of windows for
Jev and 96--98\% for Qwen2.5 7B. Macro-F1 varied across passes by at most 0.011 for Jev and 0.014
for the 7B, against 95\% intervals from resampling days about 0.2 wide. Repeat variation therefore
changes no comparison reported here.
 
\section{Discussion}\label{sec:discussion}
Table~\ref{tab:answers} gives a short answer to each research question and points to its evidence.
The subsections that follow interpret these answers.

\begin{table}[!htbp]
  \caption{Answers to the research questions.}
  \label{tab:answers}
  \small
  \begin{tabularx}{\textwidth}
    {@{}>{\raggedright\arraybackslash}p{3.4cm}>{\raggedright\arraybackslash}X
    >{\raggedright\arraybackslash}p{2.2cm}@{}}
    \toprule
    Question & Answer & Evidence \\
    \midrule
    RQ1: Which faults can training-free models diagnose?
      & Faults whose evidence one computed feature carries, such as a stuck damper or an unstable
        control loop. Not faults that need operating context. On the one unit where both could be
        compared, accuracy fell from consistency faults (L2) to context faults (L3).
      & Section~\ref{sec:res-capability}, Fig.~\ref{fig:capability} \\
    RQ2: Does computing the physics in code help?
      & Yes. Raw readings gave almost no diagnoses. Physical features raised pooled accuracy on
        consistency faults to 37\% for Jev and 46\% for Qwen2.5 72B. Named levels lowered it.
      & Section~\ref{sec:res-capability}, Fig.~\ref{fig:capability} \\
    RQ2: Do topology and point values help further?
      & Slightly for Jev, by 0.04 in shifted macro-F1, and not for Qwen2.5 72B. The gain came mainly
        from points the features lacked.
      & Sections~\ref{sec:res-capability} and \ref{sec:res-shift} \\
    RQ3: Do training-free models lose less accuracy under shift?
      & Yes. On average they lost at most 0.01 (95\% upper bound 0.10), against 0.33 for
        LightGBM. LightGBM still led or tied within a building. The training-free models led only
        across buildings, mainly because LightGBM fell below a constant ``normal'' answer.
      & Section~\ref{sec:res-shift}, Fig.~\ref{fig:shift}, Table~\ref{tab:shift} \\
    RQ3: Can their probabilities be taken at face value?
      & No. Calibration error was 0.27 or more. It did not drift under shift, and a correction
        fitted on source data mostly carried over. Detection was no better than flagging every
        window.
      & Section~\ref{sec:res-calibration}, Fig.~\ref{fig:reliability}, Table~\ref{tab:ece} \\
    \bottomrule
  \end{tabularx}
\end{table}

\subsection{What a training-free model can diagnose}\label{sec:disc-capability}
A training-free model diagnosed a fault when one physical feature, computed in code, carried its
evidence. Given raw readings, no training-free model diagnosed more than a few faults, and most of
those came from one dataset whose fault was written into a logged command. This agrees with work on
simulated sensor data from industrial equipment, where language models did better with summary
statistics than with raw readings \cite{lee2025llmfdd}. Given physical features, Jev and Qwen2.5 72B
named the faults whose evidence a single feature carries. The outdoor-air fraction and its
difference from the damper command carry a stuck damper, and the count of valve reversals per hour
carries an unstable control loop. Removing those features erased the corresponding accuracy. In
activity recognition from accelerometer data, by contrast, adding numerical features lowered Jev's
accuracy \cite{konak2026har}. A likely difference is that each feature here encodes a physical
relation that a fault breaks, so one feature could carry the diagnosis. What the models could
diagnose therefore depended more on the input (RQ2) than on the reasoning level (RQ1), and the two
questions are best answered together.

The models failed where no single feature carries the evidence. A context fault must be inferred
from the operating mode or from capacity that falls short of an expectation. On the ORNL unit
without the damper days, where only context faults remained, the training-free models scored no
better than answering ``normal'' for every window. They also failed where the evidence was missing
from the input, as for the return-fan faults until the topology input supplied the fan speed. When
they were wrong, they tended to fall back on one answer. On the shifted sets, Jev answered
``normal'' for 68\% of the ORNL winter windows, of which 14\% were fault-free, and for about 80\% of
the RP-1312 spring and summer windows, of which about 62\% were fault-free. Qwen2.5 72B instead
answered ``damper stuck'' for 59\% of the RP-1312 summer windows. Where several faults leave similar
traces, such as the leaking and stuck valves of FLEXLAB, the answers spread across them. A model
that falls back on one class scores low on the others whatever their level, which accounts for much
of the low accuracy beyond the faults a single feature carries.

The reasoning levels therefore describe how far a fault's evidence lies from the raw readings, and
the features decide how much of that distance the model must cover itself. With features, a
consistency fault (a broken physical relation between readings) or a temporal fault (a pattern over
time) came down to reading one number, while a context fault still required reasoning over several.
The decline from consistency to context faults on the ORNL unit, where both levels could be compared
on the same equipment, fits this reading. It rests on three fault types, however, and the study does
not establish a general ordering of the levels.

With the features doing most of the diagnostic work, the step left to the model was matching feature
values to the written description of each fault, and that step proved fragile. The same mixing
features that carried the ORNL damper fault drew Jev toward a damper fault on RP-1312, where
removing them raised its spring accuracy on consistency faults from 13\% to 43\%. A paraphrase of
the class descriptions with the same content produced the same jump for context faults on the ORNL
winter windows. A training-free model reads the descriptions as its only definition of each fault,
so their wording acts as a hidden part of the input.

Among the larger models, the kind of model mattered less than the input. With features, the 72B open
model named more consistency faults than Jev and fewer temporal faults, and only Jev used the fan
speed in the topology input to name the return-fan faults. Neither led at every level. The two
differed more in calibration (Section~\ref{sec:disc-trust}) and in cost and response time
(Section~\ref{sec:res-cost}). The 7B model of the same family named almost nothing, so within one
family of open models, size decided whether the model could diagnose at all. Writing the features as
named levels, as Jev's provider advises, lowered the accuracy on consistency faults for both Jev and
the 72B.

\subsection{Accuracy when conditions change}\label{sec:disc-shift}
When conditions changed, the training-free models kept their accuracy, while LightGBM lost over 40\%
of its in-domain score. This stability follows partly from design. A model that learns nothing from
the source data cannot lose what it learned, and its in-domain score was low to begin with. A user
would judge a model by its score under the new conditions, and on that score the comparison turned
on whether the building changed.

Within one building, LightGBM led or matched the training-free models in every test. On the ORNL
folds without the logged-command days it stayed ahead, while the training-free models fell to the
level of answering ``normal'' for every window. Only when the building changed did the order
reverse. Trained on the laboratory units, LightGBM labeled most fault-free windows of the occupied
building as a stuck damper, the degradation that transfer learning studies set out to repair
\cite{fan2022transfer,lei2025uda}. Jev and the 72B lost nothing there. Without the logged-command
days, however, the reversal rests on two fault days and shows LightGBM failing more than the
training-free models succeeding.

The within-building comparison favors LightGBM in a way that real buildings rarely allow. In each
seasonal test, LightGBM trained on labeled fault days of the same unit, including every fault type
it was later asked to diagnose, and a building in operation seldom has such labels. That a
supervised model trained on the same unit leads a training-free one is consistent with a heat-pump
study that compared the two directly \cite{muehlburger2025gemma}. The setting in which a
training-free model is needed, a new building without labeled faults, is the one setting in which
LightGBM collapsed. One pair of buildings is too little to rest a general claim on, and the building
shift is the test this study most needs repeated.

\subsection{Trusting the probabilities}\label{sec:disc-trust}
None of the models' stated probabilities could be taken at face value. Jev was the best calibrated
of the training-free models, but its probabilities still differed from its accuracy by about 0.27 on
the shifted sets. That is more than the within-task error of 0.17 found in an independent evaluation
of the model \cite{guo2026justaskjev}. The error was stable, however, and a stable error can be
corrected. The calibration of the training-free models did not drift when conditions changed, while
LightGBM's did. A single rescaling fitted on source data \cite{guo2017calibration} therefore carried
over to the shifted data. It brought Jev and the 72B below LightGBM in calibration error, although
for Jev it overshot into underconfidence. A probability from a training-free model can thus be
corrected once, on a labeled period from the source data, and in most of these tests the correction
held after the season, the configuration, or the building changed. LightGBM's drift left no such
option.

The fault probability, which an alarm would use, could not be trusted even after correction. Jev's
yes-or-no answer flagged most windows whether or not a fault was present, and contradicted its own
condition answer on the same windows. Taking the fault probability from the condition answer cut
false alarms sharply. Under shift, however, no model, supervised or not, detected faults better than
flagging every window, and recalibration did not change this. The recalibrated decision followed how
common faults had been in the source data, because the probabilities ranked windows too weakly to
separate them. Confident answers of ``normal'' set a further limit. On days when a fault left no
trace, the training-free models answered ``normal'' with high confidence, in line with an audit that
found Jev's confidence did not fall when the answer lay beyond what it knew
\cite{shankaranarayana2026selfknowledge}. That answer matches the data, but an operator reading the
probability cannot tell that a fault may be hidden. Whether a fault could show at all has to be
judged from the operating conditions, outside the model.

\subsection{Implications for practice}\label{sec:disc-practice}
The results point to where each kind of model fits. A unit with a labeled fault history is better
served by a supervised model trained on that history, which led or tied in every test within a
building. A unit without one, and above all a new building, is where a training-free model is
useful, since it needs no labels and kept its accuracy when the building changed.

For such a unit, the results suggest a division of work. The arithmetic of diagnosis belongs in
code, as physical features of the kind that rule sets such as APAR already compute. A training-free
model can then name the most likely of a known set of faults from those features and their written
descriptions. Its fault probability should come from the condition answer, which raised fewer false
alarms than the separate yes-or-no question, and its probabilities should be rescaled once on
whatever labeled period is available. Detection is too weak for an automatic alarm, so the answers
suit a short ranked list of candidate faults for an operator to check. Faults that need operating
context should stay with rules written for the specific system or with an engineer. Where building
data cannot leave the site, an open model run locally can take the place of Jev, provided its
probabilities are rescaled and its answer options are kept in a fixed order.

\subsection{Limitations}\label{sec:disc-limits}
The study covers air-side faults during occupied hours in four public datasets, and one version of
the decision model, jev-1.13.0, from one provider. It measures diagnostic accuracy and leaves the
energy or comfort cost of missed or false alarms to future work. The public datasets may have
appeared in the training data of the language models, which could raise their scores. A rule tree
written over the same physical features was not tested, so how much of the models' accuracy a
simpler method would match remains open.

Some effects in the design cannot be separated. Accuracy was highest where the answer set was
smallest, on the ORNL unit and Nesbitt Hall with four options each, and lowest on RP-1312 with
fifteen. The number of options is fixed by the dataset, so its effect cannot be separated from that
of the fault types and equipment. The in-domain reference set shares each day's conditions with
LightGBM's training data, which makes LightGBM's loss under shift an upper estimate. That set also
covers only the end of each day, and its mix of hours differs from that of the whole-day shifted
set, which may explain part of the small gains the training-free models showed under shift.

The answers also depend on how a request is written. The class descriptions were written once and
frozen, yet rewording them moved the ORNL accuracy on context faults by up to 30 points. Shuffling
the state fields changed up to 13\% of Jev's answers, and the order of the answer options moved the
open models' scores by 0.03--0.05, as much as the 72B's average gap to Jev. These effects are small
next to LightGBM's loss under shift, but they matter when training-free models are compared with
each other.

Measured fault data are scarce, and the day counts reflect it. Each fault intensity ran on one day,
so many intervals are wide, and the direct and temporal levels hold five days each, all from
RP-1312. One person assigned the levels, so the results by level describe these faults and do not
yet test the levels as a general scheme. Part of the accuracy on consistency faults came from faults
whose imposed value appears in a logged command. Every main result is reported with and without
those days, and a dataset built for this purpose would log the command and the actuator position
separately.
 
\section{Conclusion}\label{sec:conclusion}
This study asked where a training-free decision model can diagnose HVAC faults, what input it needs,
and how far its answers can be trusted when conditions change, using faults imposed on real
equipment in four public datasets. Table~\ref{tab:answers} answers each question. In short, such a
model diagnosed the faults whose evidence a physical feature computed in code carries, and missed
faults that need operating context. It held its accuracy and its calibration when conditions
changed, where a supervised model did not, but it started from a lower level and did not surpass the
supervised model within a building that had labeled faults. Its probabilities needed correction, and
its detection could not drive an alarm.

For practice, the results point to a division of work. Physical features are computed in code, a
training-free model ranks candidate faults with recalibrated probabilities, and an operator makes
the final call. Four extensions would test this further: more pairs of buildings with labeled
faults, datasets that log commands and actuator positions separately, a rule tree over the same
features as a baseline, and an estimate of the energy and maintenance cost of the errors that
remain. The harmonized splits, labels, and reasoning-level assignments are available from the author
on request to support that work.

\credit{\textbf{Wooyoung Jung:} Conceptualization, Methodology, Software, Formal analysis, Investigation,
Data curation, Visualization, Writing -- original draft, Writing -- review \& editing.}
\competing{The author declares no competing financial interests or personal relationships that could have
appeared to influence the work reported in this paper. The author has no relationship with TypeSafe
AI, the developer of Jev, or with Alibaba, the developer of Qwen2.5, and paid for all use of Jev
personally.}
\funding{This research did not receive any specific grant from funding agencies in the public, commercial, or
not-for-profit sectors.}
\availability{All four datasets are public: the ORNL rooftop unit \cite{granderson2023data,lbnl2022fdd}, the
RP-1312 units \cite{ghalamsiah2026data}, the FLEXLAB unit \cite{granderson2020data,lin2019oedi},
and Nesbitt Hall \cite{ghalamsiah2026data}. The splits, window labels, observability flags, and
reasoning-level and severity assignments derived in this study are available from the author on
reasonable request.}
\aiuse{During the preparation of this work the author used Claude Opus 5.5 (Anthropic) to draft and edit
the text and to assist in writing the analysis code. After using this tool, the author reviewed and
edited the content as needed and takes full responsibility for the content of the publication.}

\appendix
\renewcommand{\thetable}{\Alph{section}.\arabic{table}}
\renewcommand{\thefigure}{\Alph{section}.\arabic{figure}}
\setcounter{table}{0}
\setcounter{figure}{0}
\section{Fault types, reasoning levels, and observability}\label{app:faults}
Table~\ref{tab:faults} lists every fault type in the capability map, with its reasoning level, the
evidence that isolates it in the recorded points, and the condition under which the fault is
observable. Levels were assigned by the rule in Section~\ref{sec:data} and fixed before any model
was run. Observability was judged per window. A window counts as observable when the condition holds
for at least 15 of its 60 minutes.

{\scriptsize
\begin{xltabular}{\textwidth}
  {@{}>{\raggedright\arraybackslash}p{1.5cm}>{\raggedright\arraybackslash}p{2.3cm}c
  >{\raggedright\arraybackslash}X>{\raggedright\arraybackslash}p{3.0cm}r@{}}
  \caption{Fault types, reasoning levels, and observability conditions. Days are observable over
    total fault days. $^{\dagger}$ The imposed value is written into a logged command. OA, outdoor
    air; RA, return air; EA, exhaust air; SAT, supply-air temperature; OAT, outdoor-air
    temperature.}\label{tab:faults}\\
  \toprule
  Dataset & Fault type & Level & Evidence in the recorded points & Observable when & Days \\
  \midrule
  \endfirsthead
  \multicolumn{6}{@{}l}{Table~\ref{tab:faults}, continued}\\
  \toprule
  Dataset & Fault type & Level & Evidence in the recorded points & Observable when & Days \\
  \midrule
  \endhead
  \bottomrule
  \endlastfoot
    ORNL RTU & OA damper stuck$^{\dagger}$ & L2 & OA fraction does not follow the damper command
      & whole day; at 5\% and 10\%, only when economizing is called for & 12/16 \\
    & Economizer setpoint error & L3 & Economizer opens at the wrong outdoor temperature
      & OA between the correct and faulty changeover & 7/16 \\
    & SAT sensor bias & L3 & Supply air held at setpoint by the biased sensor; mismatch with mixed
      air and compressor stage & whole day & 16/16 \\
    \midrule
    RP-1312 & Return fan failure & L1 & Return-fan status and power drop to zero & whole day
      & 2/2 \\
    & Return fan fixed speed & L1 & Return-fan speed command held constant & whole day & 3/3 \\
    & OA damper stuck & L2 & OA flow and mixing do not follow the command
      & whole day; at 0\%, only when economizing is called for & 4/4 \\
    & OA damper leak & L2 & OA fraction above minimum with the damper at minimum
      & damper at minimum, OA--RA difference $\geq$2.8\,\textdegree C (5\,\textdegree F) & 1/4 \\
    & Cooling valve stuck & L2 & Coil temperature drop does not follow the valve command
      & fault-free twin's command $>$15 points from the stuck position & 9/9 \\
    & Cooling valve reversed & L2 & Opening the valve raises supply-air temperature & whole day
      & 1/1 \\
    & Heating valve leak & L2 & Air warms across the heating coil with the valve closed
      & whole day & 3/3 \\
    & EA damper stuck & L3 & Acts only through building pressure and return and relief flows
      & fault-free twin's command $>$15 points from the stuck position & 7/7 \\
    & Supply duct leak & L3 & Fan pressure or power high for the delivered flow & whole day & 2/2 \\
    & Heating coil fouling & L3 & Heating falls short at a given valve position
      & heating demand on the fault-free twin & 2/2 \\
    & Heating coil reduced capacity & L3 & Same as fouling & heating demand on the fault-free twin
      & 3/3 \\
    & Air filter blockage & L3 & Fan pressure or power high for the same flow & whole day & 2/2 \\
    & OAT sensor bias & L3 & Wrong economizer decisions near the changeover (the logged OAT is
      unbiased) & OA and RA within 1.7\,\textdegree C (3\,\textdegree F) & 2/2 \\
    & Unstable control & L4 & Valve, damper, or fan oscillation over time
      & economizer or mechanical cooling called for & 5/5 \\
    \midrule
    FLEXLAB & OA damper stuck & L2 & As RP-1312
      & whole day; at 10\%, only when economizing is called for & 4/4 \\
    & OA damper leak & L2 & As RP-1312 & As RP-1312 & 2/2 \\
    & Cooling valve stuck & L2 & As RP-1312 & whole day; stuck closed only with cooling demand
      & 3/4 \\
    & Cooling valve leak & L2 & Air cools across the coil with the valve closed & whole day & 3/3 \\
    & Heating valve stuck & L2 & Coil temperature rise does not follow the command
      & whole day; stuck closed only with heating demand & 5/5 \\
    & Heating valve leak & L2 & As RP-1312 & whole day & 3/3 \\
    \midrule
    Nesbitt & OA damper stuck$^{\dagger}$ & L2 & As RP-1312 & inside the documented override window
      & 6/6 \\
    & Cooling valve stuck & L2 & As RP-1312 & inside the override window & 2/2 \\
    & SAT sensor bias & L3 & As ORNL RTU & inside the override window & 2/2 \\
\end{xltabular}
}
 
\setcounter{table}{0}
\section{Feature library}\label{app:features}
The R1 representation is computed by a library of 53 features, each evaluated over one 60-minute
window from the harmonized points of Section~\ref{sec:data}. A feature that the available points
cannot support is left out of the window's payload and kept as a missing value in the LightGBM
input. Table~\ref{tab:thresholds} lists the fixed thresholds the features use, and
Table~\ref{tab:features} lists the features. The features are computed in \textdegree F, the unit of
the data given to the models, so Table~\ref{tab:features} keeps that unit. A temperature difference
of 1\,\textdegree F equals 0.56\,\textdegree C. All thresholds were set before any model was run.
The one fitted quantity, the fault-free relation between power and outdoor-air temperature, is fit
on source-domain fault-free windows only. The library has no return-fan feature, so neither R0 nor
R1 carries direct evidence of the RP-1312 return-fan faults.

\begin{table}
  \caption{Fixed thresholds used by the feature library.}
  \label{tab:thresholds}
  \footnotesize
  \begin{tabularx}{\textwidth}{@{}>{\raggedright\arraybackslash}X r@{}}
    \toprule
    Threshold & Value \\
    \midrule
    Minimum $|$OAT $-$ RAT$|$ for estimating the OA fraction &
      2.8\,\textdegree C (5\,\textdegree F) \\
    Tolerance for MAT lying between OAT and RAT & 0.6\,\textdegree C (1\,\textdegree F) \\
    Supply-fan temperature rise assumed in the APAR rules & 0.6\,\textdegree C (1\,\textdegree F) \\
    Temperature tolerance in the APAR rules & 1.1\,\textdegree C (2\,\textdegree F) \\
    Valve counts as commanded open above & 20\% \\
    Valve counts as commanded closed below & 5\% \\
    Minimum coil temperature change with the valve open & 0.6\,\textdegree C (1\,\textdegree F) \\
    Coil temperature change with the valve closed that indicates a leak &
      1.7\,\textdegree C (3\,\textdegree F) \\
    Damper counts as economizing above its minimum by & 10 points \\
    Free cooling expected (AHUs) when OAT is below RAT by & 2\,\textdegree C (3.6\,\textdegree F) \\
    Smallest command step counted as a reversal & 1 point \\
    Compressor counts as on above & 500 W \\
    Tolerance on zone temperature setpoint excursions & 0\,\textdegree C \\
    \bottomrule
  \end{tabularx}
\end{table}
{\scriptsize
\smallertable
\begin{xltabular}{\textwidth}
  {@{}>{\raggedright\arraybackslash}p{4.2cm}l>{\raggedright\arraybackslash}X
  >{\raggedright\arraybackslash}p{3.4cm}@{}}
  \caption{Physical features of the R1 representation, computed over one window. Names are shown
    exactly as they appear in the R1 text sent to the models. Each name joins the quantity and the
    statistic taken over the window, for example \texttt{oa\_frac\_est\_mean}, the mean estimated
    outdoor-air fraction. A trailing \texttt{\_frac} denotes a share of the window's rows,
    \texttt{\_std} a standard deviation, \texttt{\_maxabs} the largest absolute value, and
    \texttt{\_per\_h} a count per hour. OAT, MAT, RAT, SAT: outdoor, mixed,
    return, and supply-air temperature. ccv, hcv: cooling and heating valve commands. oad:
    outdoor-air damper command.}\label{tab:features}\\
  \toprule
  Name in the model input & Unit & Definition & Expected behavior \\
  \midrule
  \endfirsthead
  \multicolumn{4}{@{}l}{Table~\ref{tab:features}, continued}\\
  \toprule
  Name in the model input & Unit & Definition & Expected behavior \\
  \midrule
  \endhead
  \bottomrule
  \endlastfoot
  \multicolumn{4}{@{}l}{\emph{Mixing and outdoor air}}\\
  \texttt{oat\_mean} & \textdegree F
    & mean(OAT)
    & context only \\
  \texttt{rat\_mean} & \textdegree F
    & mean(RAT)
    & context only \\
  \texttt{mat\_mean} & \textdegree F
    & mean(MAT)
    & context only \\
  \texttt{oa\_rat\_diff\_mean} & \textdegree F
    & mean(OAT $-$ RAT)
    & sign tells whether outdoor air heats or cools the mix \\
  \texttt{mat\_in\_band\_frac} & --
    & share of rows with min(OAT,RAT)$-$1 $\le$ MAT $\le$ max(OAT,RAT)+1
    & drops with MAT/OAT/RAT sensor bias or duct leaks \\
  \texttt{mat\_band\_residual\_mean} & \textdegree F
    & mean distance of MAT outside that band
    & same \\
  \texttt{oa\_frac\_est\_mean} & --
    & mean (MAT$-$RAT)/(OAT$-$RAT) over rows with$|$OAT$-$RAT$|$$\ge$ 5 \textdegree{}F, clipped to
      [$-$0.2, 1.2]
    & high with OA damper stuck open or leaking; ~min with damper stuck closed \\
  \texttt{oa\_frac\_valid\_frac} & --
    & share of rows where the estimate is defined
    & low when OAT $\approx$ RAT (mild weather); the estimate is then uninformative \\
  \texttt{oa\_frac\_minus\_damper\_mean} & \%
    & mean(100$\cdot$oa\_frac\_est $-$ oad\_cmd) over valid rows
    & positive with damper leak or stuck open relative to command; large negative with stuck closed
      \\
  \texttt{oad\_cmd\_mean} & \%
    & mean(oad\_cmd)
    & constant at the stuck value for stuck-damper faults \\
  \texttt{oad\_cmd\_std} & \%
    & std(oad\_cmd)
    & zero for stuck dampers \\
  \addlinespace
  \multicolumn{4}{@{}l}{\emph{Economizer}}\\
  \texttt{oad\_above\_min\_frac} & --
    & share of rows with oad\_cmd > min + 10
    & economizer activity \\
  \texttt{econ\_expected\_frac} & --
    & share of rows where the economizer should be active (fixed changeover or OAT < RAT $-$ 3.6
      \textdegree{}F)
    & context \\
  \texttt{econ\_mismatch\_frac} & --
    & share of rows where (oad\_cmd > min + 10) disagrees with econ\_expected
    & high for economizer setpoint errors and stuck OA dampers \\
  \addlinespace
  \multicolumn{4}{@{}l}{\emph{Cooling coil}}\\
  \texttt{cc\_dt\_mean} & \textdegree F
    & mean(MAT $-$ SAT)
    & cooling delivered; near 0 with cooling valve stuck closed \\
  \texttt{ccv\_mean} & \%
    & mean(ccv)
    & saturates at 100 with valve stuck open or coil starved \\
  \texttt{cc\_open\_no\_dt\_frac} & --
    & share of rows with ccv > 20 and MAT $-$ SAT < 1
    & cooling valve stuck closed, chilled water off \\
  \texttt{cc\_uncommanded\_frac} & --
    & share of rows with ccv < 5 and MAT $-$ SAT > 3
    & cooling valve leak or stuck open \\
  \addlinespace
  \multicolumn{4}{@{}l}{\emph{Heating coil}}\\
  \texttt{hc\_dt\_mean} & \textdegree F
    & mean(SAT $-$ MAT)
    & heating delivered \\
  \texttt{hcv\_mean} & \%
    & mean(hcv)
    & -- \\
  \texttt{hc\_open\_no\_dt\_frac} & --
    & share of rows with hcv > 20 and SAT $-$ MAT < 1
    & heating valve stuck closed, coil fouled or starved \\
  \texttt{hc\_uncommanded\_frac} & --
    & share of rows with hcv < 5 and SAT $-$ MAT > 3
    & heating valve leak or stuck open \\
  \texttt{simult\_heat\_cool\_frac} & --
    & share of rows with hcv > 10 and ccv > 10
    & sequencing faults, valve leaks \\
  \addlinespace
  \multicolumn{4}{@{}l}{\emph{Supply air}}\\
  \texttt{sat\_mean} & \textdegree F
    & mean(SAT)
    & shifts with SAT sensor bias \\
  \texttt{sat\_dev\_mean} & \textdegree F
    & mean(SAT $-$ SAT setpoint)
    & nonzero with SAT sensor bias (the controller drives the biased reading to setpoint, so the
      true SAT shifts); nonzero when a coil cannot meet load \\
  \texttt{sat\_dev\_std} & \textdegree F
    & std(SAT $-$ setpoint)
    & control instability \\
  \texttt{sat\_dev\_maxabs} & \textdegree F
    & max$|$SAT $-$ setpoint$|$
    & -- \\
  \texttt{sa\_ra\_dt\_mean} & \textdegree F
    & mean(RAT $-$ SAT)
    & cooling delivered across the unit \\
  \texttt{sa\_ra\_dt\_comp\_on\_mean} & \textdegree F
    & mean(RAT $-$ SAT) while compressor 1 is on
    & low with SAT bias (+) or damper stuck open on hot days \\
  \addlinespace
  \multicolumn{4}{@{}l}{\emph{Fan and pressure}}\\
  \texttt{sf\_speed\_mean} & \%
    & mean(sf\_speed)
    & rises with filter blockage or duct leak at fixed static \\
  \texttt{sa\_sp\_dev\_mean} & in.\ w.c.
    & mean(sa\_sp $-$ sa\_sp\_sp)
    & static pressure control error \\
  \addlinespace
  \multicolumn{4}{@{}l}{\emph{Operating mode}}\\
  \texttt{mode\_heating\_frac} & --
    & hcv > 10 (or gas heat on)
    & -- \\
  \texttt{mode\_econ\_frac} & --
    & oad\_cmd > min + 10 and ccv $\le$ 10 and not heating
    & -- \\
  \texttt{mode\_econ\_cool\_frac} & --
    & oad\_cmd > min + 10 and ccv > 10
    & -- \\
  \texttt{mode\_mech\_cool\_frac} & --
    & ccv > 10 (or compressor on) and damper at minimum
    & -- \\
  \texttt{mode\_off\_frac} & --
    & fan off or unoccupied
    & -- \\
  \addlinespace
  \multicolumn{4}{@{}l}{\emph{APAR rule flags}}\\
  \texttt{apar\_r1\_frac} & --
    & heating: SAT < MAT + $\Delta$T\_sf $-$ $\varepsilon$
    & heating coil not heating \\
  \texttt{apar\_r3\_frac} & --
    & heating: hcv $\ge$ 95 and SAT < setpoint $-$ $\varepsilon$
    & heating capacity short (fouling, reduced capacity, valve stuck) \\
  \texttt{apar\_r5\_frac} & --
    & economizing:$|$SAT $-$ MAT $-$ $\Delta$T\_sf$|$> $\varepsilon$
    & coil operating while economizing \\
  \texttt{apar\_r6\_frac} & --
    & economizing:$|$OAT $-$ MAT$|$> $\varepsilon$
    & damper not fully open, leak \\
  \texttt{apar\_r13\_frac} & --
    & mechanical cooling: ccv $\ge$ 95 and SAT > setpoint + $\varepsilon$
    & cooling capacity short (valve stuck, coil fouled) \\
  \texttt{apar\_r24\_frac} & --
    & damper at minimum: oa\_frac\_est < 0.5 $\times$ min/100
    & OA damper stuck closed or ventilation short \\
  \addlinespace
  \multicolumn{4}{@{}l}{\emph{Control stability}}\\
  \texttt{ccv\_reversals\_per\_h} & 1/h
    & direction changes of ccv with$|$$\Delta$$|$> 1 pct, per hour
    & unstable valve control \\
  \texttt{hcv\_reversals\_per\_h} & 1/h
    & same for hcv
    & -- \\
  \texttt{oad\_reversals\_per\_h} & 1/h
    & same for oad\_cmd
    & unstable damper control \\
  \addlinespace
  \multicolumn{4}{@{}l}{\emph{Compressor (rooftop unit)}}\\
  \texttt{comp\_on\_frac} & --
    & share of rows with compressor 1 power > 500 W
    & cooling demand \\
  \texttt{comp2\_on\_frac} & --
    & share of rows with compressor 2 on
    & second stage use rises when SAT reads high \\
  \texttt{comp\_starts\_per\_h} & 1/h
    & rising edges of compressor-1 on state per hour
    & short cycling \\
  \texttt{comp\_min\_on\_min} & min
    & shortest on run of compressor 1 in the window
    & -- \\
  \addlinespace
  \multicolumn{4}{@{}l}{\emph{Energy and comfort}}\\
  \texttt{rtu\_power\_mean} & W
    & mean(RTU total power)
    & -- \\
  \texttt{power\_vs\_oat\_resid\_mean} & W
    & mean(power $-$ baseline(OAT)), baseline fit on source fault-free data
    & excess energy from stuck dampers, wrong economizer setpoint \\
  \texttt{zone\_temp\_mean} & \textdegree F
    & mean over zones and rows of zone temperature
    & -- \\
  \texttt{zone\_dev\_max\_mean} & \textdegree F
    & mean over rows of max over zones of the excursion outside [heating, cooling] setpoints
    & comfort impact \\
\end{xltabular}
}
 
\setcounter{table}{0}
\section{Question wording}\label{app:questions}
Each request carried a short description of the unit and its design intent, the window in one
representation, and the three questions of Table~\ref{tab:questions}. The language models also
received the instruction ``Answer with exactly one token: the label of your choice. No
explanation.'' The unit descriptions for the four datasets follow, verbatim and in the units the
models received, and Table~\ref{tab:classes} gives the description shown with every answer class.

\paragraph{Unit description, ORNL rooftop unit}
``You are diagnosing a packaged rooftop unit (RTU) serving a small office: gas heating, two-stage DX
cooling, a supply fan, and an economizer with outdoor-air and return-air dampers whose commands sum
to 100\%. Design intent in occupied mode: the outdoor-air damper sits at its 10\% minimum and
modulates open only when outdoor air is below the 50\,\textdegree F changeover (free cooling); the
supply-air setpoint is 55\,\textdegree F, but this unit normally delivers supply air a few degrees
above 55\,\textdegree F under load, so a moderate supply-air deviation on its own is not a fault.
Temperatures are in degrees\,\textdegree F, damper positions in percent open, power in W. Decide
from the physical evidence in the state, not from how often each condition occurs.''

\paragraph{Unit description, RP-1312 AHUs}
``You are diagnosing a single-duct variable-air-volume air handling unit (AHU) at a test facility in
Iowa: supply and return fans on variable-speed drives, chilled-water cooling coil, hot-water heating
coil, and outdoor, return, and exhaust air dampers with an economizer. Design intent in occupied
mode (06:00-18:00): supply air at 55\,\textdegree F (65\,\textdegree F in winter) and 1.4 in.\ w.c.
duct static; the outdoor-air damper modulates for free cooling when outdoor air is cooler than
return air, otherwise it holds its minimum. Temperatures are in degrees\,\textdegree F, damper and
valve positions in percent open, flows in cfm. Decide from the physical evidence in the state, not
from how often each condition occurs.''

\paragraph{Unit description, FLEXLAB AHU}
``You are diagnosing a single-zone air handling unit (AHU) serving one test cell: a variable-speed
supply fan, chilled-water cooling coil, hot-water heating coil, and outdoor, return, and exhaust air
dampers with an economizer. Design intent in occupied mode (06:00-18:00): the coil valves modulate
to hold the supply-air setpoint (reset between heating and cooling limits on zone demand); the
outdoor-air damper holds a 10-15\% minimum unless outdoor air is at least 3.6\,\textdegree F below
return air, when it opens for free cooling. The unit may run as constant-volume (fan fixed at 50\%)
or variable-volume (fan 10-50\%). Temperatures are in degrees\,\textdegree F, damper and valve
positions in percent open. Decide from the physical evidence in the state, not from how often each
condition occurs.''

\paragraph{Unit description, Nesbitt Hall AHUs}
``You are diagnosing a single-duct variable-air-volume air handling unit (AHU) in a commercial
building: supply and return fans, a chilled-water cooling coil, a heating coil, and outdoor, return,
and relief air dampers with an economizer. Design intent in occupied hours: the cooling coil valve
modulates to hold the supply-air setpoint; the outdoor-air damper holds its minimum (about 15\%)
unless free cooling is available. Temperatures are in degrees\,\textdegree F, damper and valve
positions in percent open. Point names differ between buildings; reason from the physical
quantities. Decide from the physical evidence in the state, not from how often each condition
occurs.''

{\scriptsize
\begin{xltabular}{\textwidth}
  {@{}>{\raggedright\arraybackslash}p{2.9cm}l>{\raggedright\arraybackslash}X
  >{\raggedright\arraybackslash}p{1.9cm}@{}}
  \caption{Answer classes, the one-word labels the language models answered with, and the
    description
    shown with each class.}\label{tab:classes}\\
  \toprule
  Class & Label & Description & Datasets \\
  \midrule
  \endfirsthead
  \multicolumn{4}{@{}l}{Table~\ref{tab:classes}, continued}\\
  \toprule
  Class & Label & Description & Datasets \\
  \midrule
  \endhead
  \bottomrule
  \endlastfoot
  \multicolumn{4}{@{}l}{\emph{ORNL rooftop unit}}\\
  Normal & normal
    & The outdoor-air damper is at or near its 10\% minimum (or modulating only when outdoor air is
      below 50\,\textdegree F), the mixed-air temperature is consistent with that damper position,
      and the supply-air temperature is in its usual range.
    & ORNL \\
  OA damper stuck & damper
    & The outdoor-air damper command is frozen at a single value (5, 10, 50, or 100\% open) that
      does not change with conditions; at 50 or 100\% the mixed-air temperature moves toward the
      outdoor-air temperature and cooling or heating energy rises. At 5 or 10\% it looks like normal
      minimum ventilation.
    & ORNL \\
  Economizer setpoint error & economizer
    & The economizer changeover is set to 42.8, 46.4, 53.6, or 57.2\,\textdegree F instead of
      50\,\textdegree F, so the damper opens for free cooling when outdoor air is above
      50\,\textdegree F, or stays at minimum when outdoor air is between the wrong setpoint and
      50\,\textdegree F. Only visible when outdoor air is in that band; otherwise it looks normal.
    & ORNL \\
  SAT sensor bias & sensor
    & The supply-air temperature sensor reads 3.6 or 7.2\,\textdegree F above or below the true
      value. The controller drives the biased reading toward 55\,\textdegree F, so the true supply
      air and the return-minus-supply temperature difference shift, and compressor staging or zone
      temperatures change, while the reported supply-air temperature itself looks on target.
    & ORNL \\
  \midrule
  \multicolumn{4}{@{}l}{\emph{Air-handling units}}\\
  Normal & normal
    & Dampers, coil valves, fans, and sensors all behave as designed for the outdoor conditions and
      the operating mode; the outdoor-air damper modulates only when economizing is appropriate, the
      mixed-air temperature lies between outdoor and return air in proportion to the damper, and
      supply air tracks its setpoint.
    & RP-1312, FLEXLAB, Nesbitt \\
  OA damper stuck & damper
    & The outdoor-air damper holds one fixed position regardless of the economizer logic or the
      command; the mixed-air temperature reflects that fixed outdoor-air share. Stuck closed means
      no ventilation air and mixed air equal to return air; stuck open means mixed air moving toward
      outdoor air and extra coil load.
    & RP-1312, FLEXLAB, Nesbitt \\
  OA damper leak & leak
    & The outdoor-air damper passes more outdoor air than its command: the estimated outdoor-air
      fraction stays above the commanded minimum when the damper should be nearly closed, with extra
      heating or cooling load in extreme weather.
    & RP-1312, FLEXLAB \\
  EA damper stuck & exhaust
    & The exhaust (relief) air damper holds a fixed position; the return and outdoor dampers still
      modulate. Building pressure and return-air flow shift, which shows as unusual return or
      outdoor airflow and fan behaviour while temperatures look near normal.
    & RP-1312 \\
  Cooling valve stuck & cooling
    & The chilled-water (cooling coil) valve holds a fixed position regardless of its command. Stuck
      closed: the command rises toward 100\% while the coil delivers no temperature drop and supply
      air runs warm. Stuck open: the coil cools even when the command is low, supply air runs cold,
      and reheat or the heating coil may fight it.
    & RP-1312, FLEXLAB, Nesbitt \\
  Cooling valve leak & seep
    & The chilled-water valve passes flow when commanded closed: a temperature drop across the
      cooling coil with the command near zero, slightly cold supply air, and extra heating to
      compensate.
    & FLEXLAB \\
  Heating valve stuck & heating
    & The hot-water (heating coil) valve holds a fixed position regardless of its command. Stuck
      closed: the command rises while supply air stays below setpoint in heating. Stuck open: supply
      air rises above setpoint with the command low, and the cooling coil may open to compensate.
    & FLEXLAB \\
  Heating valve leak & bypass
    & The hot-water valve passes flow when commanded closed: a temperature rise across the heating
      coil with the command near zero, warm supply air, and extra cooling to compensate.
    & RP-1312, FLEXLAB \\
  Return fan failure & fan
    & The return fan is stopped although it should run: return-fan status or speed is off or zero
      while the supply fan runs, return and relief airflow collapse, and the outdoor-air fraction
      rises.
    & RP-1312 \\
  Control unstable & unstable
    & A control loop hunts: the cooling valve, mixing damper, or supply fan reverses direction many
      times per hour, and supply air temperature or static pressure oscillates around its setpoint.
    & RP-1312 \\
  SAT sensor bias & sensor
    & The supply-air temperature sensor reads above or below the true value; the controller drives
      the biased reading toward setpoint so the true supply air shifts, which shows in the coil
      temperature differences and zone conditions while the reported supply air looks on target.
    & Nesbitt \\
  Return fan fixed speed & speed
    & The return fan runs at one fixed speed instead of tracking the supply fan: its speed command
      is constant while supply-fan speed varies, and return airflow and return-fan power stop
      following the supply side.
    & RP-1312 \\
  Cooling valve reverse & reverse
    & The cooling coil valve acts in reverse: opening the command reduces cooling and closing it
      increases cooling, so supply air moves the wrong way when the valve command changes.
    & RP-1312 \\
  Supply duct leak & vent
    & Supply air leaks from the duct before the zones: the supply fan runs faster or draws more
      power to hold the same static pressure and zone airflows, with no matching change at the
      coils.
    & RP-1312 \\
  Heating coil fouling & grime
    & The heating coil is fouled: for the same valve position and water temperatures it delivers
      less temperature rise, and the valve opens further than usual to hold the supply-air setpoint
      in heating.
    & RP-1312 \\
  Heating coil reduced capacity & throttled
    & The heating coil has reduced capacity (throttled water flow): the valve saturates near fully
      open while supply air falls short of its heating setpoint.
    & RP-1312 \\
  Air filter blockage & airfilter
    & The air filter is partly blocked: higher fan pressure or power for the same airflow, with coil
      behaviour unchanged.
    & RP-1312 \\
  OAT sensor bias & outdoor
    & The outdoor-air temperature used by the controller is offset from the true value, so
      economizer decisions near the changeover are wrong: free cooling is used or withheld when the
      measured outdoor and return temperatures say otherwise. No logged point carries the bias.
    & RP-1312 \\
\end{xltabular}
}

\paragraph{Example of one window}
The blocks below show what the models received for one window. The window was chosen by a fixed
rule before any answer was read: the first observable window of the first stuck-cooling-valve day in
the shifted test set of the RP-1312 summer test. It runs from 06:00 to 07:00 on 27 August 2007 on
AHU-A, with the cooling valve stuck closed. The blocks are excerpts of the state sent to Jev, in the
units the models received, with omitted parts marked ``...''. Feature units are those of
Table~\ref{tab:features}.

\medskip\noindent\begin{minipage}{\textwidth}
R0, raw readings:\par\scriptsize
\begin{verbatim}
"equipment": "AHU_A", "dataset_family": "air handling unit",
"units": "temperatures degF, damper and valve positions percent open, power W,
          flow cfm, static pressure inH2O",
"window_start": "2007-08-27 06:00", "window_end": "2007-08-27 06:59",
"season": "summer", "sampling": "5min",
"point_classes": {"oat": "Outside_Air_Temperature_Sensor",
  "mat": "Mixed_Air_Temperature_Sensor", "sat": "Supply_Air_Temperature_Sensor", ...,
  "ccv_cmd": "Valve_Command", ..., "occ": "Occupied_Mode_Status"},
"columns": ["time", "oat", "mat", "sat", "rat", "sat_sp", "oad_cmd", "rad_cmd", "ead_cmd",
  "ccv_cmd", "hcv_cmd", "sf_status", "sf_speed", "sa_sp", "sa_sp_sp", "sa_flow", "occ"],
"rows": [["06:00", 72.4, 72.5, 69.7, 73.6, 55, 6.7, 93.3, 6.7, 0, 0, 0.8, 29.9, 0.1, 1.4,
          834.9, 1],
         ["06:05", 72.5, 73.5, 70.5, 73.4, 55, 33.5, 66.5, 33.5, 0, 0, 1, 59.6, 0.6, 1.4,
          1918.7, 1],
         ...
         ["06:55", 72.8, 75.5, 78.3, 75.8, 55, 40, 60, 40, 0, 0, 1, 99, 1.4, 1.4,
          3262.5, 1]]
\end{verbatim}
\end{minipage}\par\medskip

\medskip\noindent\begin{minipage}{\textwidth}
R1, physical features (the R0 header without the sampling step and point classes, a list of
feature units, and every feature the window supports):\par\scriptsize
\begin{verbatim}
"features": {
  "oat_mean": 72.58, "rat_mean": 74.45, "mat_mean": 74.22, "sat_mean": 75.13,
  "oa_rat_diff_mean": -1.87, "mat_in_band_frac": 1.0, "mat_band_residual_mean": 0.0,
  "oa_frac_valid_frac": 0.0, "oad_cmd_mean": 36.68, "oad_cmd_std": 9.7,
  "oad_above_min_frac": 0.92, "econ_expected_frac": 0.0, "econ_mismatch_frac": 0.92,
  "cc_dt_mean": -0.91, "ccv_mean": 0.0, "cc_open_no_dt_frac": 0.0,
  "cc_uncommanded_frac": 0.08, "hc_dt_mean": 0.91, "hcv_mean": 0.0,
  "hc_open_no_dt_frac": 0.0, "hc_uncommanded_frac": 0.0, "simult_heat_cool_frac": 0.0,
  "sat_dev_mean": 20.13, "sat_dev_std": 2.8, "sat_dev_maxabs": 23.4,
  "sf_speed_mean": 79.54, "sa_sp_dev_mean": -0.27, "mode_heating_frac": 0.0,
  "mode_econ_frac": 0.92, "mode_econ_cool_frac": 0.0, "mode_mech_cool_frac": 0.0,
  "mode_off_frac": 0.02, "apar_r5_frac": 0.11, "apar_r6_frac": 0.4,
  "apar_r24_frac": 0.0, "ccv_reversals_per_h": 0.0, "hcv_reversals_per_h": 0.0,
  "oad_reversals_per_h": 0.0, "sa_ra_dt_mean": -0.68, "zone_temp_mean": 73.77,
  "zone_dev_max_mean": 2.69}
\end{verbatim}
\end{minipage}\par\medskip

\medskip\noindent\begin{minipage}{\textwidth}
R2, features and topology (R1 plus 52 Brick statements, of which 10 are shown):\par\scriptsize
\begin{verbatim}
"brick_subgraph": [
  "AHU_A isA AHU",
  "AHU_A hasPart Cooling_Coil (Cooling_Coil)",
  "AHU_A hasPart CHW_Valve (Chilled_Water_Valve)", ...,
  "AHU_A hasPoint SAT (Supply_Air_Temperature_Sensor) = 78.4",
  "AHU_A hasPoint SAT_SP (Supply_Air_Temperature_Setpoint) = 55.0", ...,
  "Return_Fan hasPoint RF_SPEED (Speed_Command) = 79.3", ...,
  "Supply_Fan hasPoint SF_SPEED (Speed_Command) = 99.1", ...,
  "OA_Damper hasPoint OAD_CMD (Damper_Position_Command) = 40.0", ...,
  "CHW_Valve hasPoint CCV_CMD (Valve_Command) = 0.0", ...,
  "AHU_A feeds VAV_1 (VAV), VAV_2 (VAV), VAV_3 (VAV), VAV_4 (VAV)"]
\end{verbatim}
\end{minipage}\par\medskip

With R0, Jev's most probable condition was a stuck cooling valve (0.51). With R1 and R2 its most
probable answer was ``normal'' (0.39 and 0.45), and the stuck valve was not among its three most
probable answers. The window was kept because the rule fixed it in advance.
 
\setcounter{table}{0}
\section{Scores under shift}\label{app:shift}
Table~\ref{tab:shift} gives the in-domain and shifted scores behind Figure~\ref{fig:shift}, with
Jev given physical features (R1) and Qwen2.5 7B added. Table~\ref{tab:ece} gives the calibration
error on each shifted test set behind the pooled values of Section~\ref{sec:res-calibration}.

\begin{table}
  \caption{Diagnosis under shift, as macro-F1 on the in-domain reference windows (In) and the
    shifted test set (Sh.). Seasons are the held-out season. FLEXLAB: VAV to CAV. Nesbitt: RP-1312
    to Nesbitt AHU-2. Days: test days in the shifted set. LightGBM receives the same input with R1
    and R2. Means are computed before rounding.}
  \label{tab:shift}
  \footnotesize
  \setlength{\tabcolsep}{2.6pt}
  \begin{tabular}{@{}llcccccccccc@{}}
    \toprule
    & & \multicolumn{4}{c}{Jev} & \multicolumn{2}{c}{LightGBM} & \multicolumn{2}{c}{Qwen2.5 72B}
      & \multicolumn{2}{c}{Qwen2.5 7B} \\
    \cmidrule(lr){3-6}\cmidrule(lr){7-8}\cmidrule(lr){9-10}\cmidrule(l){11-12}
    & & \multicolumn{2}{c}{R1} & \multicolumn{2}{c}{R2} & \multicolumn{2}{c}{R1, R2}
      & \multicolumn{2}{c}{R2} & \multicolumn{2}{c}{R2} \\
    \cmidrule(lr){3-4}\cmidrule(lr){5-6}\cmidrule(lr){7-8}\cmidrule(lr){9-10}\cmidrule(l){11-12}
    Shift test & Days & In & Sh. & In & Sh. & In & Sh. & In & Sh. & In & Sh. \\
    \midrule
    ORNL, fall & 20 & 0.29 & 0.35 & 0.32 & 0.35 & 0.68 & 0.51 & 0.30 & 0.35 & 0.14 & 0.25 \\
    ORNL, spring & 14 & 0.33 & 0.31 & 0.36 & 0.31 & 0.74 & 0.61 & 0.36 & 0.25 & 0.17 & 0.18 \\
    ORNL, summer & 14 & 0.36 & 0.24 & 0.38 & 0.24 & 0.75 & 0.49 & 0.36 & 0.24 & 0.19 & 0.18 \\
    ORNL, winter & 14 & 0.33 & 0.28 & 0.32 & 0.36 & 0.73 & 0.73 & 0.30 & 0.34 & 0.21 & 0.10 \\
    RP-1312, spring & 19 & 0.40 & 0.39 & 0.65 & 0.57 & 0.72 & 0.48 & 0.41 & 0.37 & 0.13 & 0.12 \\
    RP-1312, summer & 18 & 0.29 & 0.21 & 0.42 & 0.37 & 0.72 & 0.35 & 0.33 & 0.19 & 0.11 & 0.10 \\
    RP-1312, winter & 12 & 0.17 & 0.29 & 0.20 & 0.29 & 0.85 & 0.25 & 0.24 & 0.40 & 0.16 & 0.15 \\
    FLEXLAB & 13 & 0.32 & 0.22 & 0.35 & 0.16 & 0.59 & 0.25 & 0.13 & 0.12 & 0.09 & 0.02 \\
    Nesbitt & 30 & 0.42 & 0.68 & 0.46 & 0.72 & 0.99 & 0.12 & 0.42 & 0.49 & 0.30 & 0.33 \\
    \midrule
    Mean & & 0.33 & 0.33 & 0.39 & 0.37 & 0.75 & 0.42 & 0.32 & 0.31 & 0.17 & 0.16 \\
    \bottomrule
  \end{tabular}
\end{table}

\begin{table}
  \caption{Calibration error (ECE) on each shifted test set, as stated (Raw) and after recalibration
    fitted on the in-domain windows (Recal.). Pooled: all windows of the nine tests together.}
  \label{tab:ece}
  \small
  \begin{tabular}{@{}lcccccc@{}}
    \toprule
    & \multicolumn{2}{c}{Jev, R2} & \multicolumn{2}{c}{LightGBM, R1}
      & \multicolumn{2}{c}{Qwen2.5 72B, R2} \\
    \cmidrule(lr){2-3}\cmidrule(lr){4-5}\cmidrule(l){6-7}
    Shift test & Raw & Recal. & Raw & Recal. & Raw & Recal. \\
    \midrule
    ORNL, fall & 0.32 & 0.20 & 0.32 & 0.21 & 0.46 & 0.19 \\
    ORNL, spring & 0.39 & 0.18 & 0.20 & 0.11 & 0.66 & 0.20 \\
    ORNL, summer & 0.59 & 0.14 & 0.35 & 0.23 & 0.70 & 0.22 \\
    ORNL, winter & 0.31 & 0.19 & 0.11 & 0.07 & 0.51 & 0.13 \\
    RP-1312, spring & 0.08 & 0.09 & 0.12 & 0.09 & 0.23 & 0.11 \\
    RP-1312, summer & 0.10 & 0.30 & 0.23 & 0.14 & 0.37 & 0.18 \\
    RP-1312, winter & 0.19 & 0.13 & 0.62 & 0.51 & 0.13 & 0.23 \\
    FLEXLAB & 0.36 & 0.11 & 0.50 & 0.29 & 0.52 & 0.06 \\
    Nesbitt & 0.19 & 0.19 & 0.76 & 0.76 & 0.25 & 0.10 \\
    \midrule
    Pooled & 0.28 & 0.17 & 0.36 & 0.27 & 0.43 & 0.16 \\
    \bottomrule
  \end{tabular}
\end{table}

\end{document}